\documentclass[10pt,twocolumn,letterpaper]{article}
\usepackage[pagenumbers]{cvpr} % To force page numbers, e.g. for an arXiv version

\usepackage{graphicx}
\usepackage{amsmath}
\usepackage{amssymb}
\usepackage{booktabs}
\usepackage{kotex}
\usepackage{times}
\usepackage{epsfig}
\usepackage{color}
\usepackage{adjustbox}
\usepackage{algorithm}
\usepackage{algorithmic}
\usepackage{multirow}
\usepackage{lipsum}
\usepackage{enumitem}
\usepackage{animate}
\usepackage[accsupp]{axessibility}  % Improves PDF readability for those with disabilities.

\newcommand\norm[1]{\lVert#1\rVert}

\definecolor{cvprblue}{rgb}{0.21,0.49,0.74}
\usepackage[pagebackref,breaklinks,colorlinks,allcolors=cvprblue]{hyperref}

\def\paperID{4} % *** Enter the Paper ID here
\def\confName{CVPR}
\def\confYear{2025}

\title{ARC-NeRF: Area Ray Casting for Broader Unseen View Coverage\\in Few-shot Object Rendering}

\author{Seunghyeon Seo$^{1}$~~~
Yeonjin Chang$^{1}$~~~
Jayeon Yoo$^{1}$~~~
Seungwoo Lee$^{1}$~~~
Hojun Lee$^{1,2}$~~~
Nojun Kwak$^{1\S}$
\smallskip
\\
$^{1}$Seoul National University~~~
$^{2}$Xperty Corp.\\
{\tt\small \{zzzlssh, yjean8315, jayeon.yoo, seungwoo.lee, hojun815, nojunk\}@snu.ac.kr}
}

\begin{document}
\twocolumn[{
\renewcommand\twocolumn[1][]{#1}
\maketitle
\vspace{-2.0mm}
\begin{tabular}{p{0.15\textwidth}p{0.14\textwidth}p{0.14\textwidth}p{0.14\textwidth}p{0.14\textwidth}p{0.14\textwidth}}
     \centering\scriptsize Mip-NeRF~\cite{barron2021mip} & \centering\scriptsize RegNeRF~\cite{niemeyer2022regnerf} & \centering\scriptsize  FlipNeRF~\cite{seo2023flipnerf} & \centering\scriptsize FreeNeRF~\cite{yang2023freenerf} & \centering\scriptsize \textbf{ARC-NeRF} & \centering\scriptsize Ground Truth
\end{tabular}
\includegraphics[width=\textwidth]{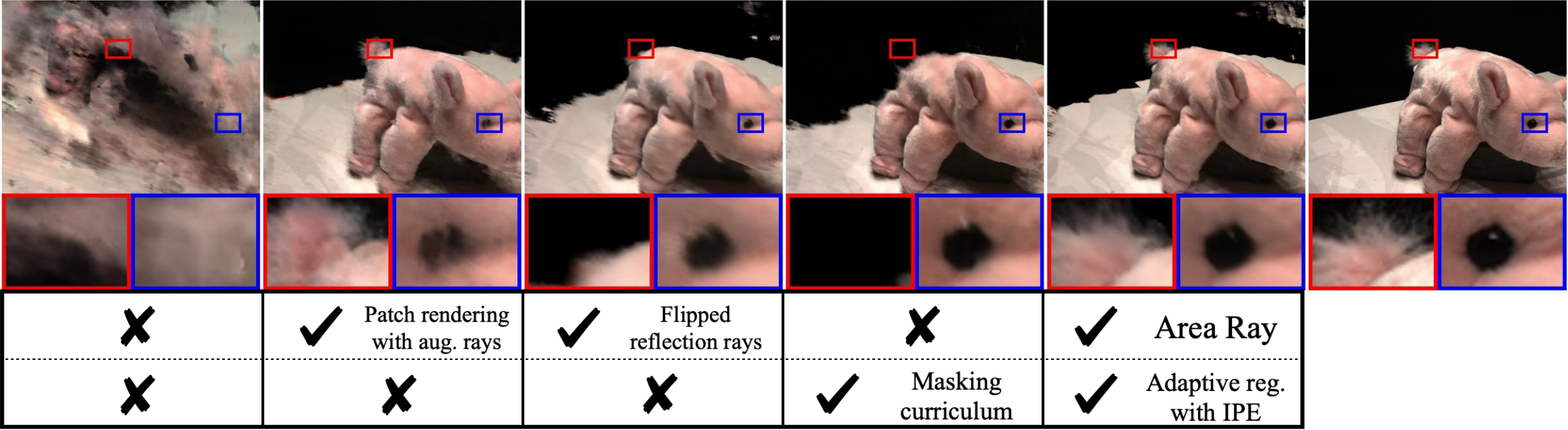}
\vspace{-5.5mm}
\captionof{figure}{\textbf{Comparison with other baselines.}
Our ARC-NeRF achieves superior quality of fine details and textures to other state-of-the-art methods by using our Area Rays equipped with an adaptive high-frequency regularization.
The last two rows of the figure indicate the types of ray augmentation schemes and frequency regularization methods, respectively.}
\label{fig:intro}
\vspace{4mm}
}]
\begin{abstract}
{\let\thefootnote\relax\footnotetext{{$^{\S}$Corresponding author.}}}
Recent advancements in the Neural Radiance Field (NeRF) have enhanced its capabilities for novel view synthesis, yet its reliance on dense multi-view training images poses a practical challenge, often leading to artifacts and a lack of fine object details.
Addressing this, we propose ARC-NeRF, an effective regularization-based approach with a novel Area Ray Casting strategy.
While the previous ray augmentation methods are limited to covering only a single unseen view per extra ray, our proposed Area Ray covers a broader range of unseen views with just a single ray and enables an adaptive high-frequency regularization based on target pixel photo-consistency.
Moreover, we propose luminance consistency regularization, which enhances the consistency of relative luminance between the original and Area Ray, leading to more accurate object textures.
The relative luminance, as a free lunch extra data easily derived from RGB images, can be effectively utilized in few-shot scenarios where available training data is limited.
Our ARC-NeRF outperforms its baseline and achieves competitive results on multiple benchmarks with sharply rendered fine details.
\vspace{-4mm}
\end{abstract}    
\section{Introduction}
\label{sec:intro}

In recent years, the Neural Radiance Field (NeRF)~\cite{mildenhall2021nerf} has emerged as a dominant paradigm in the domain of novel view synthesis, owing to its outstanding capability to produce high-quality rendered images and its inherently simplistic architectural design.
Numerous advancements have been introduced to the NeRF~\cite{lin2023magic3d, chang2024fast, liu2023zero, metzer2023latent, jiang2022neuman, weng2022humannerf, li2023dynibar, li2022neural}, taking its performance to the next level.
However, its dependency on a dense set of multi-view training images still exists as a critical challenge for practical application, resulting in floating artifacts and missing fine details in object textures.

Two primary approaches have emerged as mainstream methodologies for the few-shot novel view synthesis: \textit{pre-training} and \textit{regularization}.
The pre-training methods~\cite{chen2021mvsnerf, chibane2021stereo, wang2021ibrnet, liu2022neural, jang2021codenerf, li2021mine, rematas2021sharf, trevithick2021grf, johari2022geonerf, yu2021pixelnerf} necessitate extensive datasets comprising diverse scenes captured from multiple viewpoints, facilitating the infusion of prior knowledge for 3D geometry during the pre-training phase.
However, it is highly expensive to collect large-scale datasets necessary for pre-training.

In contrast, the regularization methods~\cite{niemeyer2022regnerf, seo2023mixnerf, kim2022infonerf, jain2021putting, roessle2022dense, deng2022depth, kwak2023geconerf, seo2023flipnerf, yang2023freenerf, truong2023sparf, song2023harnessing, wang2023sparsenerf, somraj2023simplenerf, wynn2023diffusionerf, xiong2023sparsegs} are optimized per scene, using extra training assets, such as rays generated from unseen views~\cite{seo2023flipnerf, niemeyer2022regnerf, kim2022infonerf, kwak2023geconerf}, pseudo-depth maps~\cite{deng2022depth, roessle2022dense, truong2023sparf, wang2023sparsenerf}, readily available off-the-shelf models~\cite{jain2021putting, niemeyer2022regnerf, wynn2023diffusionerf}, and so on.
Moreover, these methods may directly apply regularization techniques to the internal components of a NeRF framework, \eg the density outputs of ray samples~\cite{seo2023mixnerf} and the high-frequency segments stemming from positional encoding~\cite{yang2023freenerf, song2023harnessing}.
Although these methods have achieved promising results, they often rely on extra training resources, which may not be consistently accessible.
Also, the existing frequency regularization technique requires the range of masked frequency and the duration of masking phase, both of which should be set manually.

To address the problem, we propose \textit{ARC-NeRF}, which utilizes a novel Area Ray casting strategy as an effective ray augmentation scheme.
We cast an Area Ray as a bundle of multiple additional rays, by featurizing the 3D Gaussian space of conical frustum using the Integrated Positional Encoding (IPE)~\cite{barron2021mip}.
This enables us to cover a broader area of unseen views as an additional training resource, enhancing the efficiency of augmented rays, whereas previous ray augmentation methods are limited to covering only a single unseen view per additional ray.

Furthermore, by applying IPE to the Area Ray, we can adaptively apply high-frequency regularization based on the photo-consistency of the pixels where the original input ray and the newly generated Area Ray are cast.
Since the photo-consistency varies depending on the angle between the original and Area Ray, which determines the radius and variance of Area Ray, the high-frequency components of Area Ray samples are adaptively regulated based on the photo-consistency.
It allows us to eliminate the hand-crafted aspects of the existing frequency regularization scheme and capture fine details better while preventing from overfitting on the high-frequency.

Additionally, we propose luminance map, which serves as free lunch additional data that can be easily derived from RGB images, as an effective additional training resource in few-shot scenarios with limited training data, and introduce \textit{luminance consistency regularization} with an auxiliary luminance estimation task.
It enhances the consistency of estimated luminance between the original ray and its corresponding Area Ray, allowing for more accurate capture of delicate object textures with fine details as shown in~\cref{fig:intro}.
Our main contributions are as follows:
\begin{itemize}[noitemsep,topsep=0pt,parsep=0pt,partopsep=0pt, leftmargin=*]
\item We propose a novel ray augmentation strategy, \ie casting an Area Ray as a bundle of additional rays to cover a broader range of unseen views, which is effective for training NeRF with only a set of sparse inputs.
\item By marrying the IPE with our Area Ray, we can adaptively regularize the high-frequency of additional input samples based on the photo-consistency, leading to improved detail in rendering under the few-shot scenario.
\item We propose a luminance map, easily derived from RGB images as simple but effective free lunch data, and introduce luminance consistency regularization through an auxiliary luminance estimation task, further enhancing the rendering quality.
\item Our ARC-NeRF outperforms its baseline, delivering competitive results across different datasets and scenarios with sharp fine details in object textures.
\end{itemize}
\section{Related Works}
\label{sec:related_works}
%-------------------------------------------------------------------------
\subsection{Neural Radiance Fields}
\label{subsec:nerf}

The advent of NeRF~\cite{mildenhall2021nerf} has facilitated a paradigm shift in novel view synthesis, achieving remarkable rendering quality.
The NeRF employs a Multilayer Perceptron (MLP) to represent a scene, correlating 3D coordinates and viewing directions with corresponding color and volumetric density attributes, and then generates a novel view by volume rendering techniques.
While NeRF's versatility has been demonstrated across various applications, \eg the 3D object generation~\cite{lin2023magic3d, poole2022dreamfusion, liu2023zero, metzer2023latent}, capturing human performances~\cite{xu2021h, zhao2022humannerf, su2021nerf, jiang2022neuman, weng2022humannerf}, dynamic videos~\cite{gao2021dynamic, li2023dynibar, li2022neural, tretschk2021non, pumarola2021d}, and so on, its efficacy is notably hindered by a fundamental constraint: the reliance on a dense set of training images.
This requirement poses significant challenges, particularly in scenarios restricted to a scant number of views, thus hampering its practical application.
In addressing this critical bottleneck, our work focuses on improving NeRF's performance under the constraints of limited input.

%-------------------------------------------------------------------------
\subsection{Few-shot Neural Rendering}
\label{subsec:fewshot}

In the emerging field of few-shot novel view synthesis, two principal strategies are prevalent: \textit{pre-training} and \textit{regularization}.
The pre-training methods~\cite{chen2021mvsnerf, chibane2021stereo, wang2021ibrnet, liu2022neural, jang2021codenerf, li2021mine, rematas2021sharf, trevithick2021grf, johari2022geonerf, yu2021pixelnerf} rely on large-scale datasets of multi-view scenes to infuse a NeRF model with a 3D geometry prior, often followed by fine-tuning on specific target scenes.
In contrast, the regularization methods~\cite{niemeyer2022regnerf, seo2023mixnerf, kim2022infonerf, jain2021putting, roessle2022dense, deng2022depth, kwak2023geconerf, seo2023flipnerf, yang2023freenerf, truong2023sparf, song2023harnessing, somraj2023simplenerf, wynn2023diffusionerf, wang2023sparsenerf, xiong2023sparsegs} are optimized to individual scenes by leveraging auxiliary training aids, \eg pseudo-depth maps~\cite{deng2022depth, roessle2022dense, truong2023sparf, wang2023sparsenerf}, augmented training rays~\cite{seo2023flipnerf, niemeyer2022regnerf, kim2022infonerf, kwak2023geconerf}, semantic consistencies~\cite{jain2021putting}, and so on for additional guidance.

Among them, the ray augmentation methods~\cite{seo2023flipnerf, niemeyer2022regnerf, kim2022infonerf, kwak2023geconerf} directly address the lack of training data by generating additional rays for training, effectively tackling data limitations.
However, since the number of augmented rays directly increases the training cost, generating a large number of additional rays is practically challenging.
Therefore, the existing ray augmentation methods have limitations in fully utilizing available unseen views.
A notable technique within this approach, FlipNeRF~\cite{seo2023flipnerf} exploits flipped reflection rays as additional training data, yielding promising outcomes.
However, since only a single reflection ray is used per original ray for an unseen view, it still does not leverage the potentially effective additional rays that exist between them from unseen views.

Recently, the studies tackling the issue of overfitting to high-frequency in the few-shot scenario~\cite{yang2023freenerf, song2023harnessing} have emerged.
FreeNeRF~\cite{yang2023freenerf} pioneered a frequency regularization approach to adjust the visible frequency range based on the training time steps.
Despite its meaningful observation and promising results, it requires a manually designed masking phase and strong prior specific to a dataset, showing a lack of fine details depending on the scenes' structures due to the heuristically regulated high-frequency.

In this work, we propose a novel Area Ray casting strategy that uses a featurized bundle of multiple rays to cover not just a single unseen view, but a broad range of unseen views.
Furthermore, by applying IPE to featurize the Area Ray, we can regulate the high-frequency of the Area Ray adaptively based on the target pixel photo-consistency between the original input ray and the corresponding Area Ray, leading to the improved quality with clear fine details.

%-------------------------------------------------------------------------
\subsection{Ray Parameterization of NeRF}
\label{subsec:ray_param}

Since the original NeRF casts an infinitesimally narrow ray with point sampling, there has been a line of research exploring the effective ray shapes for higher-quality rendering, \ie method to parameterize the ray~\cite{verbin2022ref, barron2021mip, barron2022mip, isaac2023exact}.
Among them, mip-NeRF~\cite{barron2021mip} proposed a cone tracing strategy, in which each conical frustum area of the samples is featurized by the IPE, to tackle the aliasing problem.
Mip-NeRF 360~\cite{barron2022mip} utilized contracted scene representation, which is an extended version of the mip-NeRF's parameterization to unbounded scenes.
Recently, Exact-NeRF~\cite{isaac2023exact} introduced a pyramidal parameterization method by using an analytically precise encoding technique instead of the IPE's 3D Gaussian approximation, leading to performance improvement for unbounded scenes.
In this paper, we propose a novel Area Ray casting strategy as an effective ray augmentation scheme for few-shot NeRF, which we conceptualize as a featurized bundle of multiple rays, covering wider unseen view areas and enabling an adaptive frequency regularization by combining with IPE.
To the best of our knowledge, our ARC-NeRF is the first to tackle the few-shot novel view synthesis task through the lens of ray parameterization.
\section{Methods}
\label{sec:methods}

\begin{figure}[!t]
\centering
\includegraphics[width=\linewidth]{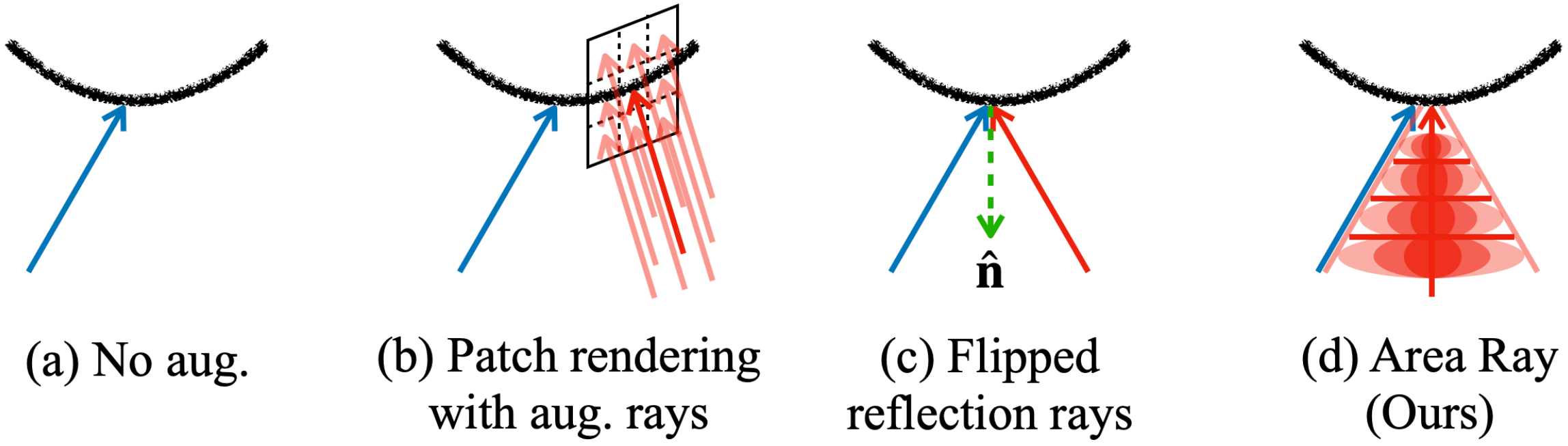}
\vspace{-5mm}
\caption{\textbf{Comparative overview of ray augmentation techniques.}
        In contrast to other augmentation strategies ((b), (c)), where each additional ray corresponds to a single unseen view, our Area Ray (d) encompasses a wider area of unseen views, thereby boosting the augmentation's efficacy.
        Technically, the rays are spanned over the object surface as used in the typical NeRF's volume rendering technique, but we simplify the illustration for easy comparison between the methods.
        Kindly refer to \cref{fig:hg_generation} for a more detailed description.
        The {\color{blue}blue} and {\color{red}red} rays indicate the original and augmented ray, respectively.
        }
\vspace{-4mm}
\label{fig:aug_comparison}
\end{figure}

In this work, we propose an effective regularization method for few-shot novel view synthesis with an Area Ray casting strategy.
Our ARC-NeRF is built upon FlipNeRF, which utilizes flipped reflection rays as the augmented rays (\cref{subsec:preliminaries}).
Instead of the flipped reflection rays, we generate a set of Area Rays whose conical frustums are featurized by IPE and cast them toward the identical target pixels as additional training rays, adaptively regulating their high-frequency components (\cref{subsec:hourglass}).
Furthermore, by taking advantage of the luminance map, \ie free lunch data that can be easily derived from RGB images, we propose the \textit{luminance consistency regularization} to enhance the consistency of the estimated relative luminances between the original ray and its corresponding Area Ray, leading to further performance improvement (\cref{subsec:luminance}).
A brief comparison with different ray augmentation schemes is shown in~\cref{fig:aug_comparison}.

%-------------------------------------------------------------------------
\subsection{Preliminaries}
\label{subsec:preliminaries}
%-------------------------------------------------------------------------
\noindent \textbf{Mip-NeRF. \ }
The mip-NeRF~\cite{barron2021mip}, which casts cones instead of rays to alleviate aliasing, represents a 3D scene by mapping 3D coordinates $\mathbf{x} = (x, y, z)$ and viewing directions $(\theta, \phi)$ to the colors $\mathbf{c} = (r, g, b)$ and volumetric densities $\tau$:
\begin{equation}
  f(\gamma^{*}(\mathbf{x}), \gamma(\mathbf{d})) \to (\mathbf{c}, \tau),
  \label{eq:nerf}
\end{equation}
where $f(\cdot, \cdot)$, $\gamma(\cdot)$, $\gamma^{*}(\cdot)$ and $\mathbf{d}$ denote an MLP-based network, original positional encoding, IPE, and 3D Cartesian unit vector practically used as an input viewing direction, respectively.
Compared to the original NeRF which utilizes point-sampling, mip-NeRF featurizes the conical frustum area $\mathrm{F}(\mathbf{d}, \mathbf{o}, \dot{\rho}, t_i, t_{i+1})$ by an approximated IPE, where $\mathbf{o}$ and $\dot{\rho}$ indicate the camera origin and base radius of a cone, respectively:
\begin{align}
\begin{split}
        \gamma^{*}&(\mathbf{x}) = \gamma^{*}(\boldsymbol{\mu}, \boldsymbol{\Sigma}) = \mathbb{E}_{x \sim \mathcal{N}(\mathbf{P} \boldsymbol\mu, \mathbf{P} \mathbf{\Sigma} \mathbf{P}^\top)} \left[ \gamma(\mathbf{x}) \right] \\
        &= 
        \begin{bmatrix}
        \sin(\mathbf{P} \boldsymbol{\mu}) \circ \exp(-0.5 \text{diag}(\mathbf{P} \mathbf{\Sigma} \mathbf{P}^\top)) \\
        \cos(\mathbf{P} \boldsymbol{\mu}) \circ \exp(-0.5 \text{diag}(\mathbf{P} \mathbf{\Sigma} \mathbf{P}^\top))
        \end{bmatrix}\,.
\label{eq:ipe}
\end{split}
\end{align}
Here, $\boldsymbol{\mu}$, $\boldsymbol{\Sigma}$, $\mathbf{P}$, and $\circ$ denote the mean and covariance of the conical frustum approximated with multivariate Gaussians, the positional encoding basis matrix, and the element-wise multiplication, respectively.
We kindly refer to \cite{barron2021mip} for more technical details.

Then each target pixel is rendered by alpha compositing the output colors and densities along the input cone $\mathbf{r}(t) = \mathbf{o} + t\mathbf{\bar{d}}$, where $\mathbf{\bar{d}} = \norm{\mathbf{\bar{d}}}_2 \cdot \mathbf{d}$.
In practice, the volume rendering integrals are approximated using a quadrature rule~\cite{mildenhall2021nerf} as follows:
\begin{equation}
\mathbf{\hat{c}(r)} = \sum_{i=1}^{N}w_i\mathbf{c}_i = \sum_{i=1}^{N}T_i(1-\exp(-\tau_i\delta_i))\mathbf{c}_i,
  \label{eq:alpha_composition}
\end{equation}
where $T_i = \exp ( -\sum_{j=1}^{i-1}\tau_j\delta_j )$.
Note that $w_i$, $\delta_i$ and $N$ indicate the alpha blending weight, the interval between adjacent samples and the number of samples, respectively.
Finally, the radiance field is trained by minimizing the mean squared errors (MSE) between the rendered pixels and GTs:
\begin{equation}
  \mathcal{L}_\text{MSE} = \sum_{\mathbf{r} \in \mathcal{R}} || \mathbf{\hat{c}(r)} - \mathbf{c_\text{GT}(r)} ||^2_2\,,
  \label{eq:mse}
\end{equation}
where $\mathcal{R}$ is a batch of inputs.

%-------------------------------------------------------------------------
\noindent \textbf{FlipNeRF. \ }
Implemented upon mip-NeRF, the FlipNeRF~\cite{seo2023flipnerf} utilizes a set of flipped reflection rays $\mathbf{r}'$ as additional training rays, which are derived from the original ray directions $\mathbf{\bar{d}}$ and the estimated normal vectors $\mathbf{\hat{n}} = \sum_{i=1}^{N} w_{i}\mathbf{n}_{i}$ as follows:
\begin{equation}
\mathbf{r}'(t) = \mathbf{o}' + t\mathbf{d}',
  \label{eq:flip_ref_ray}
\end{equation}
where $\mathbf{d}' = 2(\mathbf{\bar{d}} \cdot \hat{\mathbf{n}})\hat{\mathbf{n}} - \mathbf{\bar{d}}$ and $\mathbf{o}' = \mathbf{p}_s - t_s\mathbf{d}'$ indicate the flipped reflection direction and the newly set camera origin derived from the object's estimated surface $\mathbf{p}_s = \mathbf{o} + t_s\mathbf{\bar{d}}$, respectively.
Note that $t_s$ is the distance to the estimated object surface.
By utilizing the GT pixel values of original rays to provide relevant supervisory signals for additional rays, the flipped reflection rays can be optimized effectively with filtering the ineffective rays whose angle with the corresponding original rays are over the masking threshold $\psi=90^\circ$.
However, despite the countless potential candidates for effective additional rays in unseen view areas under the masking threshold, only the flipped reflection ray is utilized, leading to suboptimal training efficiency.

Our ARC-NeRF is built upon FlipNeRF while achieving better rendering quality with more fine details compared to FlipNeRF thanks to our proposed Area Ray casting.
Furthermore, we conceptualize the Area Ray as a bundle of multiple rays to cover a broad unseen view region, more optimally utilizing ray augmentation.

%-------------------------------------------------------------------------
\subsection{Area Ray Casting}
\label{subsec:hourglass}
\begin{figure}[!t]
\begin{subfigure}[b]{\linewidth}
\centering
\includegraphics[width=0.85\linewidth]{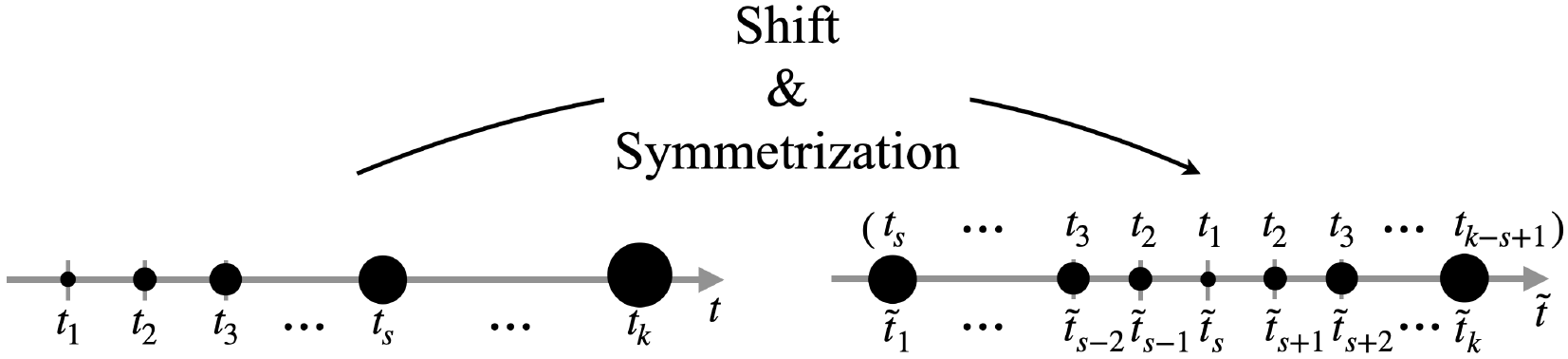}
\subcaption{Metric distance reparameterization}
\label{fig:t-reparameterization}
\vspace{2mm}
\end{subfigure}
\begin{subfigure}[b]{\linewidth}
\centering
\includegraphics[width=\linewidth]{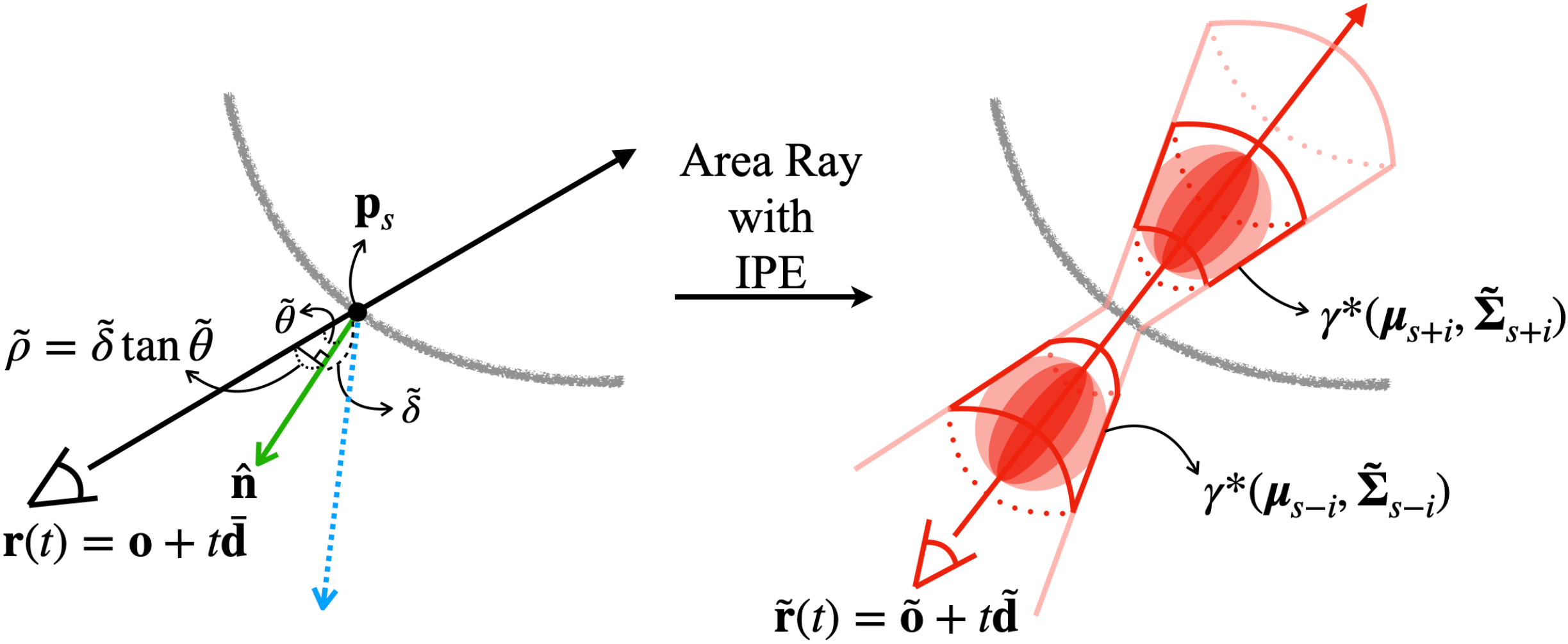}
\subcaption{Area Ray featurization by IPE}
\label{fig:hg_featurization}
\end{subfigure}
\vspace{-7mm}
\caption{\textbf{Area Ray generation process.}
        (a) First, to derive $\tilde{\sigma}^2_\rho$, we reparameterize the original metric distance $t$ as $\tilde{t}$.
        We shift $t_1$, \ie starting point of a ray along a z-axis, to the estimated object surface $\mathbf{p}_s$, so that our proposed Area Ray is symmetrically constructed around $\mathbf{p}_s$, \ie $\tilde{t}_{s-i} = \tilde{t}_{s+i}$ and it leads to $\boldsymbol{\tilde{\Sigma}}_{s-i} = \boldsymbol{\tilde{\Sigma}}_{s+i}$.
        Note that $\tilde{t}$ is used only for $\tilde{\sigma}^2_\rho$.
        (b) Using the trigonometric function, we compute the base radius of Area Ray $\tilde{\rho}$.
        As a result, our proposed Area Ray $\mathbf{\tilde{r}}$ is featurized to cover the unseen view area between the original ray and its reflection ray.
        The {\color{blue}blue} dotted ray denotes the reflection ray of $\mathbf{r}$.
        }
\vspace{-4mm}
\label{fig:hg_generation}
\end{figure}

We propose an Area Ray as an effective additional training ray for few-shot NeRF.
As shown in~\cref{fig:aug_comparison}, compared to the existing ray augmentation schemes where a resulting augmented ray corresponds to an unseen view, our proposed Area Ray covers the area of continuous unseen views by IPE presented in~\cref{eq:ipe}, providng more efficient extra training resources.

First, we reparameterize the metric distance $t \in [t_{near},$ $t_{far}]$ as $\tilde{t}$ to derive the variance $\tilde{\sigma}^2_\rho$, which is perpendicular to the Area Ray (\cref{fig:t-reparameterization}).
We shift $t$ so that the value of starting point $t_1$ is located on the distance to the estimated object surface $t_s$, resulting in $\tilde{t}_s = t_1$.
And then we set $[t_1, t_{s-1}]$ values to be symmetric with respect to $t_s$.

Next, as shown in~\cref{fig:hg_featurization}, we derive a base radius of the Area Ray $\tilde{\rho}$ from the angle $\tilde{\theta}$ between $\mathbf{r}$ and $\mathbf{\hat{n}}$ using the trigonometric function as follows:
\begin{equation}
\tilde{\rho} = \tilde{\delta}\tan{\tilde{\theta}},
  \label{eq:hg_base_radius}
\end{equation}
where $\tilde{\delta} = 1 - \tilde{t}_s$ so that $\tilde{\rho}$ is obtained from the sample located on $\tilde{t} = 1$ following~\cite{barron2021mip}.
However, directly employing the obtained $\tilde{\rho}$ in IPE results in significantly large $\tilde{\sigma}^2_\rho$, leading to over-regularization of high-frequency components for the samples along an Area Ray.
Thus, we adjust the scale of $\tilde{\rho}$ to $[0, 1]$ to contract $\tilde{\rho}$ with the large $\tilde{\theta}$ value into a proper range, while leaving the one with small $\tilde{\theta}$ affected little as follows:
\begin{equation}
\tilde{\rho} = \exp{(-1 / (\tilde{\delta}\tan{\tilde{\theta}}))}.
  \label{eq:hg_base_radius_normalization}
\end{equation}

And then, $\tilde{\sigma}^2_\rho$ is derived from $\tilde{t}$ and $\tilde{\rho}$ to featurize the conical frustums of Area Ray as multivariate Gaussian by simply replacing the original metric distance $t$, which is used in mip-NeRF, with $\tilde{t}$ as follows:
\begin{equation}
\tilde{\sigma}^2_\rho = \tilde{\rho}^2\left( \frac{\tilde{t}^2_\mu}{4} + \frac{5\tilde{t}^2_\delta}{12} - \frac{4\tilde{t}^4_\delta}{15(3\tilde{t}^2_\mu + \tilde{t}^2_\delta)} \right),
  \label{eq:hg_variance}
\end{equation}
where $\tilde{t}_\delta$ and $\tilde{t}_\mu$ denote a half-width and mid-point of adjacent $\tilde{t}$ values.
Note that we use the same $\mu_t$ and $\sigma^2_t$ for the mean and variance along the Area Ray as mip-NeRF.

Finally, we generate an Area Ray $\mathbf{\tilde{r}}(t) = \mathbf{\tilde{o}} + t\mathbf{\tilde{d}}$, where $\mathbf{\tilde{d}} = -\mathbf{\hat{n}}$ and $\mathbf{\tilde{o}} = \mathbf{p}_s - t_s\mathbf{\tilde{d}}$, so that the Area Ray is cast from the newly set camera origin $\mathbf{\tilde{o}}$, which has the same distance from $\mathbf{p}_s$ as the original ray, covering the unseen view area between the original ray and the corresponding reflection ray around the axis of $\mathbf{\hat{n}}$.
Implemented upon FlipNeRF, our ARC-NeRF is optimized with the same training losses while using our proposed Area Rays as an additional training batch instead of flipped reflection rays.

Since the target pixel photo-consistency between $\mathbf{r}$ and $\mathbf{\tilde{r}}$ varies depending on the angle $\tilde{\theta}$ between them, which is used to derive the base radius of Area Ray $\tilde{\rho}$ and the x/y-axis variance $\tilde{\sigma}^2_\rho$, the high-frequency spectrum of samples along $\mathbf{\tilde{r}}$ are regulated adaptively based on the photo-consistency, \ie the less photo-consistent the target pixel is, the more regulated the samples' high-frequency components are.
It eliminates the need for manually designing a frequency spectrum to regularize during the training phase and effectively adjusts the amount of high-frequency detail retained in additional training samples for learning fine details.

%------------------------------------------------------------------------
\subsection{Luminance Consistency Regularization}
\label{subsec:luminance}

We propose the luminance map as an effective additional training resource for few-shot scenarios with limited data, providing 'free lunch' information easily derived from RGB images, and introduce \textit{luminance consistency regularization}.

For simplicity, we use a relative luminance value, which is normalized as $[0, 1]$, and derive the GT relative luminance $y_\text{GT}$ of a target pixel as follows:
\begin{equation}
y_\text{GT} = \sum_{\bar{c}}^{\{\bar{r}, \bar{g}, \bar{b}\}} \lambda_{\bar{c}} \bar{c},
  \label{eq:gt_luminance}
\end{equation}
where $\bar{c} = c_\text{GT}^{2.2}$ indicates a linear rgb component converted from the gamma-compressed one by applying a simple power curve~\cite{poynton2012digital}.
We set the linear coefficients $\lambda_\mathbf{\bar{c}} = \left( \lambda_{\bar{r}}, \lambda_{\bar{g}}, \lambda_{\bar{b}} \right)$ as $0.2126$, $0.7152$, and $0.0722$, respectively, considering that green light, as the predominant element of luminance, contributes the most to human light perception, with blue light being the least contributing one~\cite{reinhard2023photographic, poynton1996technical}.

In addition to the existing outputs, our ARC-NeRF estimates the luminance $y$ as additional outputs per sample along a ray
and renders the final luminance $\hat{y}$ by volume rendering as follows:
\begin{equation}
\hat{y}(\mathbf{r}) = \sum_{i=1}^{N}w_i y_i,
\label{eq:lum_rendering}
\end{equation}
where $y_i \in [0, 1]$ is the estimated relative luminance of the $i$-th sample along a ray $\mathbf{r}$.
The estimated luminance map is trained to minimize MSE:
\begin{equation}
  l_\text{lum.} = \sum_{\mathbf{r} \in \mathcal{R}} || \hat{y}(\mathbf{r}) - y_\text{GT}(\mathbf{r}) ||^2_2.
  \label{eq:lum_loss}
\end{equation}
The luminance map estimated from the Area Ray is also optimized using the same $y_\text{GT}$ by minimizing the corresponding MSE loss $\tilde{l}_\text{lum.}$, resulting in our proposed training loss $\mathcal{L}_\text{lum.} = \eta l_\text{lum.} + \tilde{\eta} \tilde{l}_\text{lum.}$, where $\eta$ and $\tilde{\eta}$ are balancing weights.

By optimizing the volumetric rendered luminance, we can provide extra supervisory signals, whose GT luminances are obtained from GT pixel values without any burdensome process, to the blending weights $w$, which are directly related to the estimated depth map of NeRF.
As a result, it leads to performance improvement under the few-shot scenario with more accurate object textures.
Further details about the total loss and architecture are provided in the supplementary material.

\section{Experiments}
\label{sec:experiments}
%-------------------------------------------------------------------------
\subsection{Experimental Details}
\label{subsec:exp_details}

\noindent \textbf{Datasets and metrics. \ }
We evaluate our ARC-NeRF and other baselines on three representative datasets for novel view synthesis: Realistic Synthetic 360$^\circ$~\cite{mildenhall2021nerf}, DTU~\cite{jensen2014large}, and Shiny Blender~\cite{verbin2022ref}, focusing on the ability to capture fine details of objects.
Realistic Synthetic 360$^\circ$ has 8 synthetic scenes, each containing 400 multi-view images with a white background.
We use 4 and 8 views to train our ARC-NeRF, and conduct an analysis of the frequency regularization effect and the effect of luminance estimation within the 4-view setting, using the first 4 and 8 images from the training set for a fair comparison, following the protocol in~\cite{seo2023mixnerf, seo2023flipnerf}.
On the DTU dataset, featuring multi-view images with objects against a white table and black background, we compare ARC-NeRF against other methods across 3/6/9-view scenarios and perform an ablation study in the 3-view context, following~\cite{yu2021pixelnerf}.
Furthermore, we evaluate our method on Shiny Blender, which comprises 6 synthetic glossy objects.

For quantitative evaluation, we use the metrics such as PSNR, SSIM~\cite{wang2004image}, LPIPS~\cite{zhang2018unreasonable}, and the geometric average~\cite{barron2021mip}, each averaged across all scenes within each dataset.
Note that we use the masked evaluation metrics to avoid the background bias and focus on the target objects.

\begin{figure}[!t]
\centering
\includegraphics[width=\linewidth]{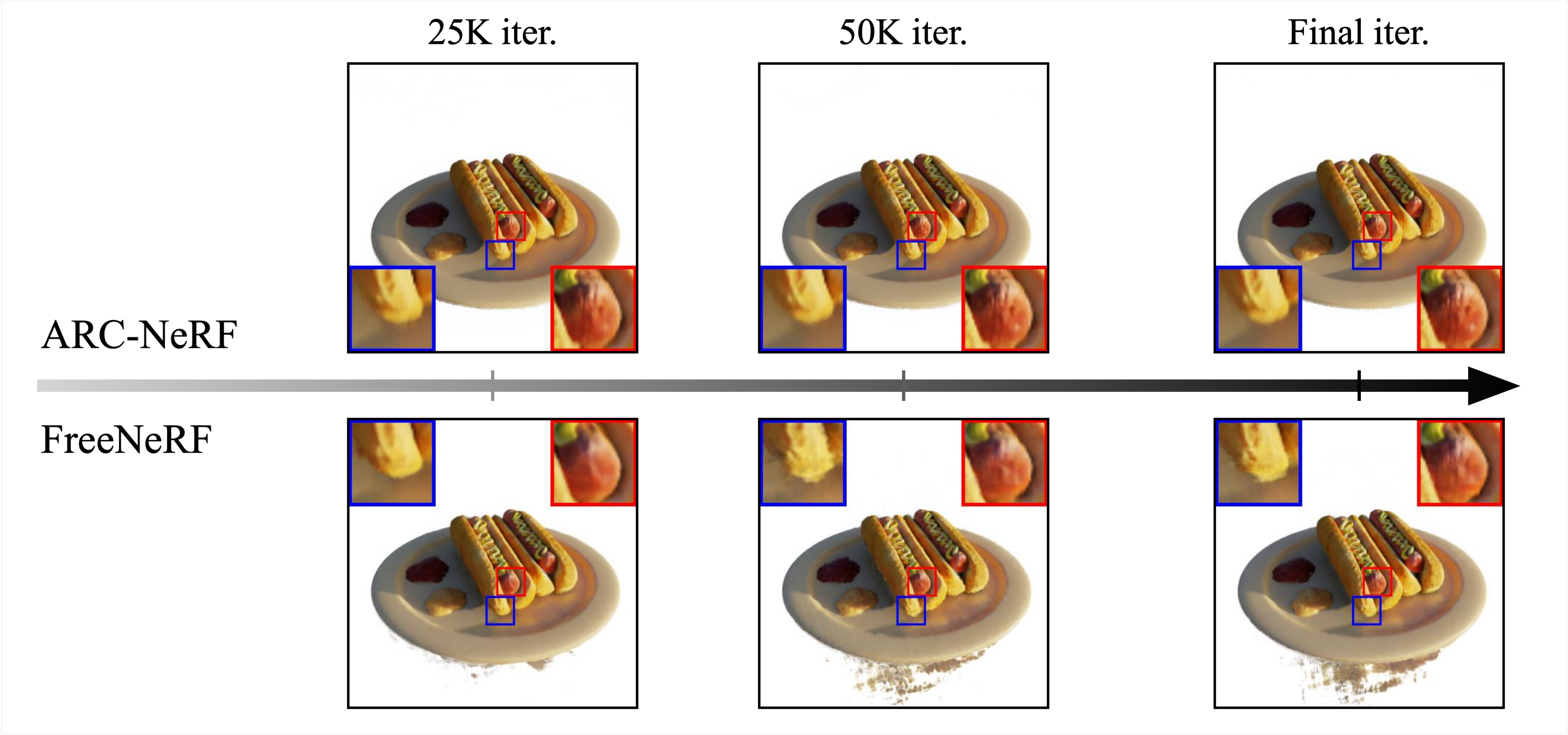}
\caption{\textbf{Comparison of fine details against FreeNeRF over the training phase.}
        Compared to FreeNeRF which forcibly masks the high-frequency spectrum in the early training phase, ours adaptively regularizes the high-frequency components of additional ray samples based on the target pixel photo-consistency (\ie the angle between the original ray and Area Ray) during the whole training process.
        As a result, our ARC-NeRF already achieves sharper fine details at 25K iteration than the fully trained FreeNeRF.
        }
\vspace{-5mm}
\label{fig:freq_anal}
\end{figure}

\noindent \textbf{Baselines. \ }
We compare our ARC-NeRF against the state-of-the-art (SOTA) regularization methods~\cite{jain2021putting, kim2022infonerf, niemeyer2022regnerf, seo2023mixnerf, yang2023freenerf, seo2023flipnerf} and the original mip-NeRF~\cite{barron2021mip} on Realistic Synthetic 360$^\circ$.
We also evaluate against pre-training~\cite{chen2021mvsnerf, chibane2021stereo, yu2021pixelnerf} and more regularization methods~\cite{somraj2023simplenerf, wynn2023diffusionerf, wang2023sparsenerf, xiong2023sparsegs}, including 3DGS-based approaches~\cite{kerbl20233d,xiong2023sparsegs} which have become another mainstream, on the DTU dataset.
On Shiny Blender, our ARC-NeRF is compared with FlipNeRF~\cite{seo2023flipnerf} and FreeNeRF~\cite{yang2023freenerf}, designed for ray augmentation and high-frequency regularization, respectively, as well as Ref-NeRF~\cite{verbin2022ref} addressing the ray parameterization for view-dependent effects.
The pre-training methods utilize DTU for pre-training, whereas regularization approaches and mip-NeRF are optimized per scene.
Note that we report the results of other methods from~\cite{niemeyer2022regnerf, seo2023mixnerf, seo2023flipnerf}, which outperformed the results from the corresponding original paper by modified training curriculum~\cite{niemeyer2022regnerf} and used the same training views to ensure a fair comparison~\cite{seo2023mixnerf, seo2023flipnerf}.
Kindly refer to our supplementary material for more details.

%-------------------------------------------------------------------------
\subsection{Analysis of ARC-NeRF}
\label{subsec:analysis}

\noindent \textbf{Frequency regularization effect of Area Ray. \ }
\cref{fig:freq_anal} demonstrates the comparison between our ARC-NeRF and FreeNeRF, which is specialized for frequency regularization.
Ours renders sharper fine details than FreeNeRF from the earlier training phase and already achieves competitive rendering quality only after 25K training iterations compared to the fully trained FreeNeRF.
Although FreeNeRF prevents overfitting by forcibly masking most of high-frequency components during early training, it results in non-satisfactory rendering outcomes that fail to capture fine details in the end.
In contrast, ours adaptively regularizes the high-frequency components of the Area Ray samples based on the target pixel photo-consistency, preventing overfitting and enabling sharper rendering.

\begin{figure}[!t]
\centering
\includegraphics[width=\linewidth]{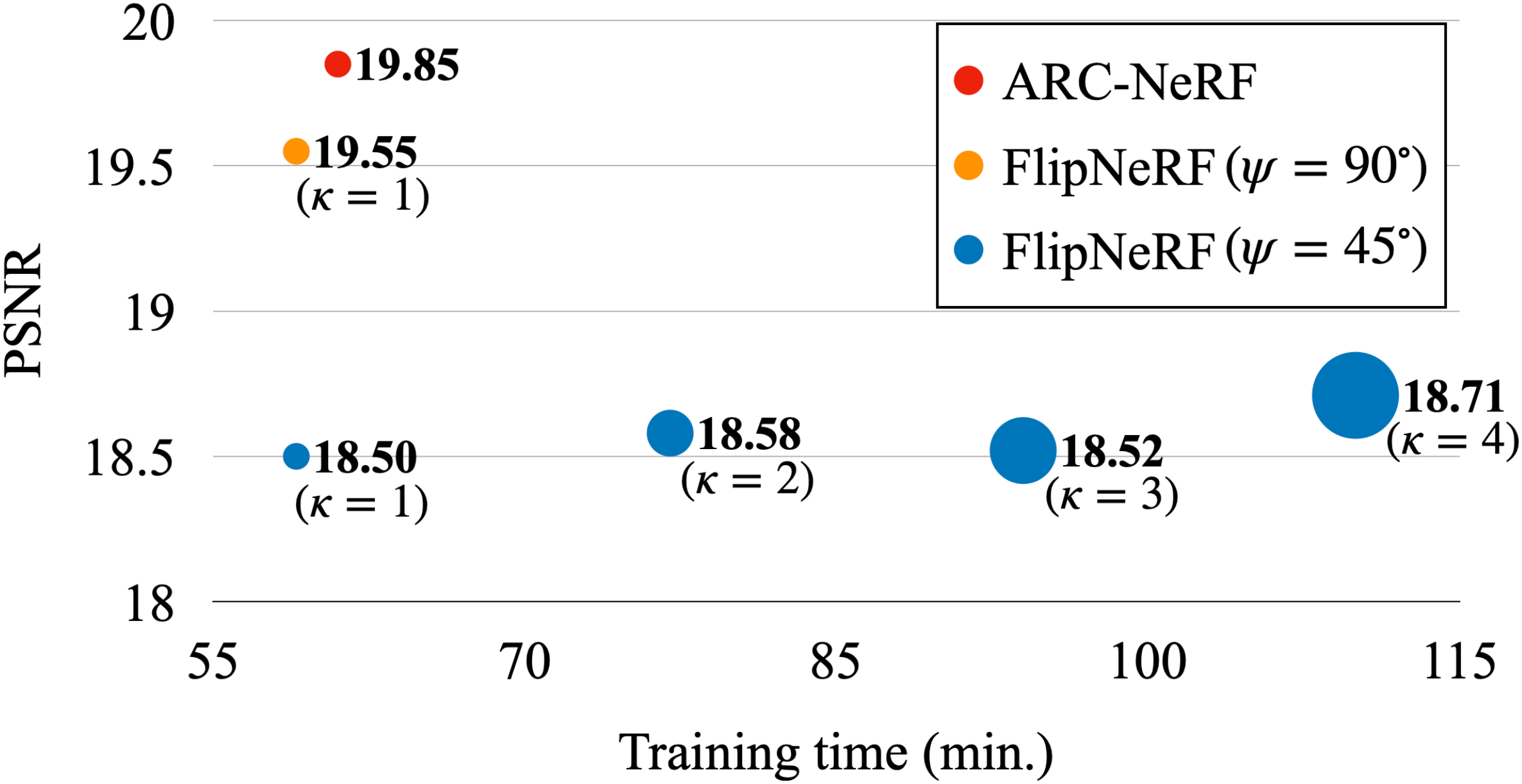}
\vspace{-7mm}
\caption{\textbf{Comparison of our ARC-NeRF against multicasting strategy on DTU 3-view.}
        Our ARC-NeRF outperforms FlipNeRF in all scenarios by a large margin.
        The training time per scene is measured using the same GPU, iterations, and batch size.
        The size of circles is proportional to $\kappa$, \ie the number of augmented rays per original ray.
        }
\label{fig:multicasting_exp}
\end{figure}

\begin{figure*}[!t]
\centering
\begin{tabular}{p{0.02\textwidth}p{0.128\textwidth}p{0.128\textwidth}p{0.128\textwidth}p{0.128\textwidth}p{0.128\textwidth}p{0.128\textwidth}p{0.03\textwidth}}
     & \centering\scriptsize RegNeRF~\cite{niemeyer2022regnerf} & \centering\scriptsize FlipNeRF~\cite{seo2023flipnerf} & \centering\scriptsize  FreeNeRF~\cite{yang2023freenerf} & \centering\scriptsize \textbf{ARC-NeRF} & \centering\scriptsize \textbf{ARC-NeRF$^\dagger$} & \centering\scriptsize Ground Truth &
    \end{tabular}
\includegraphics[width=0.92\linewidth]{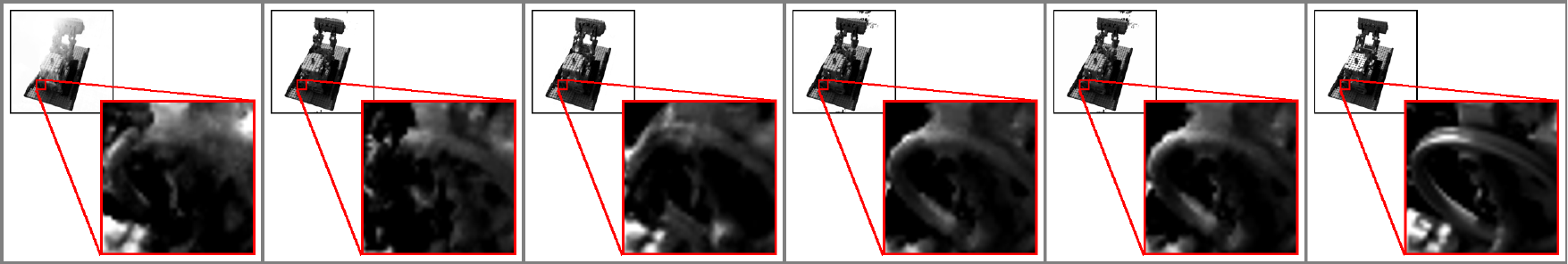}
\vspace{-2mm}
\caption{\textbf{Qualitative results of relative luminance maps.}
        Our ARC-NeRF renders clearer relative luminance map than other baselines.
        With our explicitly estimated luminance $\hat{y}$, ours$^\dagger$ captures the detailed luminance on the handle more accurately.
        The {\color{red}red} insets are visualized with doubled contrast for ease of comparison.
        }
\vspace{-4mm}
\label{fig:lum_comparison}
\end{figure*}

\noindent \textbf{Effectiveness of Area Ray as a bundle of rays. \ }
In~\cref{fig:multicasting_exp}, we demonstrate the effectiveness of Area Ray, which covers a broad range of unseen views, compared to FlipNeRF with a different number of augmented rays, \ie the multicasting strategy.
Based on FlipNeRF, we cast $\kappa$ extra rays per original ray $\mathbf{r}$, consisting of flipped reflection ray $\mathbf{r}'$ and additional rays which are evenly spaced between $\mathbf{r}$ and its corresponding $\mathbf{r}'$, \ie within the area of our Area Ray, toward the same target pixel.
When $\kappa=1$, which amounts to the vanilla FlipNeRF, our ARC-NeRF outperforms FlipNeRF by a large margin, showing the efficacy of Area Ray.
However, exploiting $\kappa > 1$ augmented rays, fails to achieve meaningful performance improvement.
This suggests that our single Area Ray casting, equipped with adaptive frequency regularization, is more effective than multicasting strategy over the same unseen area.
Moreover, for a training efficiency, simply increasing the number of augmented rays in the multicasting strategy leads to longer training time due to the additional ray processing.
However, our ARC-NeRF maintains training efficiency while improving performance by featurizing continuous unseen views with a single Area Ray.
Note that we set the masking threshold $\psi=45^\circ$ identically for a fair comparison\footnote{By using multicasting strategy with $\psi=90^\circ$, which is original masking threshold for FlipNeRF, the FlipNeRF suffers from severe training instability and fails to achieve comparable results.
We conjecture that using an excessive number of augmented rays at the initial training phase, when the model is not sufficiently trained to generate valid additional rays, leads to the failure.}.

\begin{table}[!t]
\centering
\resizebox{\linewidth}{!}{
\begin{tabular}{l|c|c|c|c}
\toprule
& {PSNR $\uparrow$} & {SSIM $\uparrow$} & {LPIPS $\downarrow$} & {Average $\downarrow$} \\
\midrule
RegNeRF~\cite{niemeyer2022regnerf} & 5.71 & 0.780 & 0.291 & 0.368 \\
FlipNeRF~\cite{seo2023flipnerf} & 16.49 & 0.878 & 0.080 & 0.092 \\
FreeNeRF~\cite{yang2023freenerf} & 15.63 & 0.869 & 0.091 & 0.113 \\
\textbf{ARC-NeRF}  & \underline{16.87} & \underline{0.886} & \underline{0.074} & \underline{0.088} \\
\textbf{ARC-NeRF$^\dagger$}  & \textbf{17.18} & \textbf{0.891} & \textbf{0.062} & \textbf{0.079} \\
\bottomrule
\end{tabular}
}
\vspace{-2mm}
\caption{
\textbf{Comparison of relative luminance map.}
$\dagger$ indicates the explicitly estimated luminance map $\hat{y}$ while others compute lumniance based on estimated RGB values.
}
\label{tab:lum_exp}
\vspace{-5mm}
\end{table}

\noindent \textbf{Luminance estimation as an auxiliary task. \ }
We show the quantitative and qualitative results of our proposed luminance estimation in~\cref{tab:lum_exp} and \cref{fig:lum_comparison}, respectively.
Since other baselines do not estimate the relative luminance $\hat{y}$ explicitly, their estimated luminance maps $\hat{y}^{\mathbf{c}}$ are derived from the estimated colors using~\cref{eq:gt_luminance} and we report our results of both $\hat{y}^{\mathbf{c}}$ and $\hat{y}$ for a fair comparison.
Ours outperforms other representative methods based on the ray augmentation and frequency regularization.
As illustrated in~\cref{fig:lum_comparison}, our ARC-NeRF renders much clearer relative luminance maps than other baselines.
Since the resulting relative luminances and pixel values share the identical blending weights $w$ for volumetric rendering, we are able to provide extra supervisory signals to the blending weights and effectively regularize them by the luminance estimation task.

%-------------------------------------------------------------------------
\noindent \textbf{Ablation study. \ }
\cref{tab:ablation} shows the ablation study of our ARC-NeRF.
By replacing the flipped reflection rays of FlipNeRF with our Area Ray (1), it achieves performance improvement across most of the metrics, especially on SSIM and LPIPS by a large margin.
However, it achieves rather degenerate results with only $\mathcal{L}_\text{lum.}$ (2).
Since the flipped reflection rays are technically based on the mirror reflection, which assumes that the pixel's luminance varies depending on its viewpoint, our $\mathcal{L}_\text{lum.}$ goes against the physical concept of FlipNeRF training framework, leading to the performance drop.
When we optimize our ARC-NeRF with our proposed $\mathcal{L}_\text{lum.}$ together ((1) $\rightarrow$ (3)), we can further enhance the rendering quality with clear fine details, especially on PSNR metric.
Additional experiment of the masking threshold for Area Ray is provided in our supplementary material.

\begin{table}[!t]
\centering
\resizebox{\linewidth}{!}{
\begin{tabular}{c|cc|c|c|c|c}
\toprule
& Area Ray & $\mathcal{L}_\text{lum.}$ & {PSNR $\uparrow$} & {SSIM $\uparrow$} & {LPIPS $\downarrow$} & {Avg. $\downarrow$} \\
\midrule
FlipNeRF~\cite{seo2023flipnerf} & & & \underline{19.55} & 0.767 & 0.180 & 0.101 \\
\midrule
(1) & \checkmark & & 19.51 & \textbf{0.774} & \underline{0.147} & \underline{0.097} \\
(2) & & \checkmark & 18.44 & 0.747 & 0.201 & 0.119 \\
(3) & \checkmark & \checkmark & \textbf{19.85} & \underline{0.773} & \textbf{0.146} & \textbf{0.096} \\
\bottomrule
\end{tabular}
}
\vspace{-2mm}
\caption{
\textbf{Ablation study.}
Thanks to the proposed Area Ray and $\mathcal{L}_\text{lum.}$, our ARC-NeRF achieves superior performance to its baseline FlipNeRF across all metrics.
}
\label{tab:ablation}
\vspace{-6mm}
\end{table}

\begin{table}[!t]
\centering
\resizebox{\linewidth}{!}{
\begin{tabular}{l|cc|cc|cc|cc}
\toprule
  \multirow{2}{*}{} & \multicolumn{2}{c}{PSNR $\uparrow$} & \multicolumn{2}{c}{SSIM $\uparrow$} & \multicolumn{2}{c}{LPIPS $\downarrow$} & \multicolumn{2}{c}{Avg. $\downarrow$}  \\
  & 4-view & 8-view & 4-view & 8-view & 4-view & 8-view & 4-view & 8-view \\ \midrule
Mip-NeRF~\cite{barron2021mip} & 8.70 & 13.31 & 0.792 & 0.848 & 0.250 & 0.176 & 0.285 & 0.188 \\
\midrule
DietNeRF~\cite{jain2021putting} & 10.86 & 16.08 & 0.814 & 0.870 & 0.194 & 0.113 & 0.223 & 0.123 \\
InfoNeRF~\cite{kim2022infonerf} & 13.65 & 16.74 & 0.834 & 0.865 & 0.134 & 0.094 & 0.139 & 0.095 \\
RegNeRF~\cite{niemeyer2022regnerf} & 7.24 & 13.47 & 0.795 & 0.856 & 0.292 & 0.158 & 0.318 & 0.177 \\
MixNeRF~\cite{seo2023mixnerf} & \cellcolor{yellow!50}{16.13} & \cellcolor{yellow!50}{19.31} & \cellcolor{yellow!50}{0.863} & \cellcolor{yellow!50}{0.902} & \cellcolor{yellow!50}{0.099} & \cellcolor{yellow!50}{0.058} & \cellcolor{yellow!50}{0.101} & \cellcolor{yellow!50}{0.065} \\
FreeNeRF~\cite{yang2023freenerf} & 15.71 & 18.99 & 0.857 & 0.894 & 0.103 & 0.064 & 0.114 & 0.072 \\
FlipNeRF~\cite{seo2023flipnerf} & \cellcolor{orange!50}{16.47} & \cellcolor{orange!50}{19.54} & \cellcolor{orange!50}{0.866} & \cellcolor{orange!50}{0.903} & \cellcolor{orange!50}{0.091} & \cellcolor{orange!50}{0.057} & \cellcolor{orange!50}{0.095} & \cellcolor{orange!50}{0.062} \\
\textbf{ARC-NeRF } & \cellcolor{red!50}{\textbf{16.86}} & \cellcolor{red!50}{\textbf{20.29}} & \cellcolor{red!50}{\textbf{0.873}} & \cellcolor{red!50}{\textbf{0.910}} & \cellcolor{red!50}{\textbf{0.084}} & \cellcolor{red!50}{\textbf{0.052}} & \cellcolor{red!50}{\textbf{0.091}} & \cellcolor{red!50}{\textbf{0.057}} \\
\bottomrule
\end{tabular}
}
\vspace{-2mm}
\caption{
    \textbf{Quantitative results on Realistic Synthetic 360$^\circ$.}
    ARC-NeRF outperforms others across all the scenarios and metrics.
    }
\label{tab:blender}
\vspace{-8mm}
\end{table}

%-------------------------------------------------------------------------
\subsection{Comparison with other Baselines}
\label{subsec:main_exp}
\begin{figure*}[!t]
\centering
\begin{tabular}
{p{0.02\textwidth}p{0.128\textwidth}p{0.128\textwidth}p{0.128\textwidth}p{0.128\textwidth}p{0.128\textwidth}p{0.128\textwidth}p{0.03\textwidth}}
     & \centering\scriptsize mip-NeRF~\cite{barron2021mip} & \centering\scriptsize MixNeRF~\cite{seo2023mixnerf} & \centering\scriptsize  FreeNeRF~\cite{yang2023freenerf} & \centering\scriptsize FlipNeRF~\cite{seo2023flipnerf} & \centering\scriptsize \textbf{ARC-NeRF} & \centering\scriptsize Ground Truth &
    \end{tabular}
\begin{subfigure}[b]{0.92\textwidth}
         \centering
        \includegraphics[width=\linewidth]{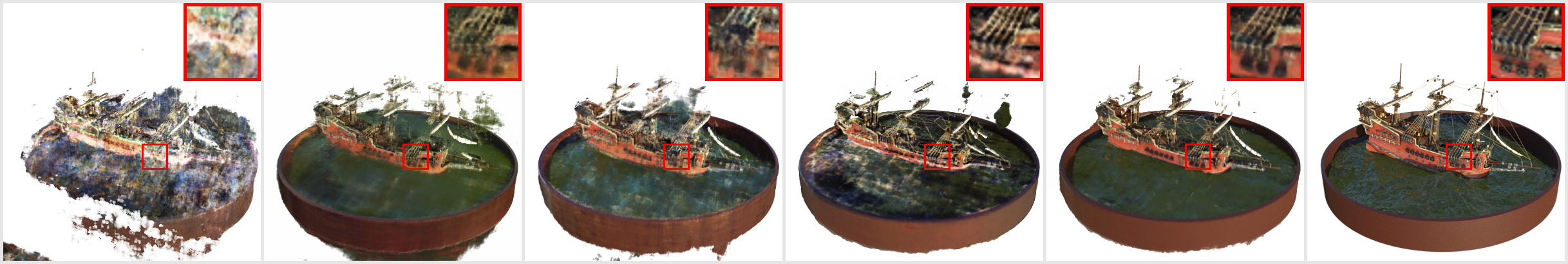}
         \caption{\scriptsize Realistic Synthetic 360$^\circ$ 4-view}
         \label{fig:blender}
     \end{subfigure}
\begin{tabular}
{p{0.02\textwidth}p{0.128\textwidth}p{0.128\textwidth}p{0.128\textwidth}p{0.128\textwidth}p{0.128\textwidth}p{0.128\textwidth}p{0.03\textwidth}}
     & \centering\scriptsize mip-NeRF~\cite{barron2021mip} & \centering\scriptsize RegNeRF~\cite{niemeyer2022regnerf} & \centering\scriptsize  FreeNeRF$^\ddagger$~\cite{yang2023freenerf} & \centering\scriptsize FlipNeRF~\cite{seo2023flipnerf} & \centering\scriptsize \textbf{ARC-NeRF} & \centering\scriptsize Ground Truth &
    \end{tabular}
\begin{subfigure}[b]{0.92\textwidth}
         \centering
        \includegraphics[width=\linewidth]{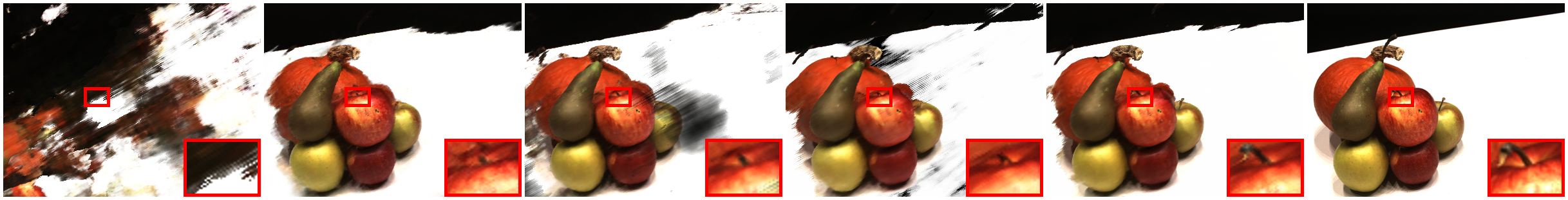}
         \caption{\scriptsize DTU 3-view}
         \label{fig:dtu}
     \end{subfigure}
\begin{tabular}{p{0.04\textwidth}p{0.122\textwidth}p{0.122\textwidth}p{0.12\textwidth}p{0.12\textwidth}p{0.122\textwidth}p{0.04\textwidth}}
     & \centering\scriptsize Ref-NeRF~\cite{verbin2022ref} & \centering\scriptsize  FreeNeRF~\cite{yang2023freenerf} & \centering\scriptsize FlipNeRF~\cite{seo2023flipnerf} & \centering\scriptsize \textbf{ARC-NeRF} & \centering\scriptsize Ground Truth &
    \end{tabular}
\begin{subfigure}[b]{0.92\textwidth}
         \centering
        \includegraphics[width=0.8\linewidth]{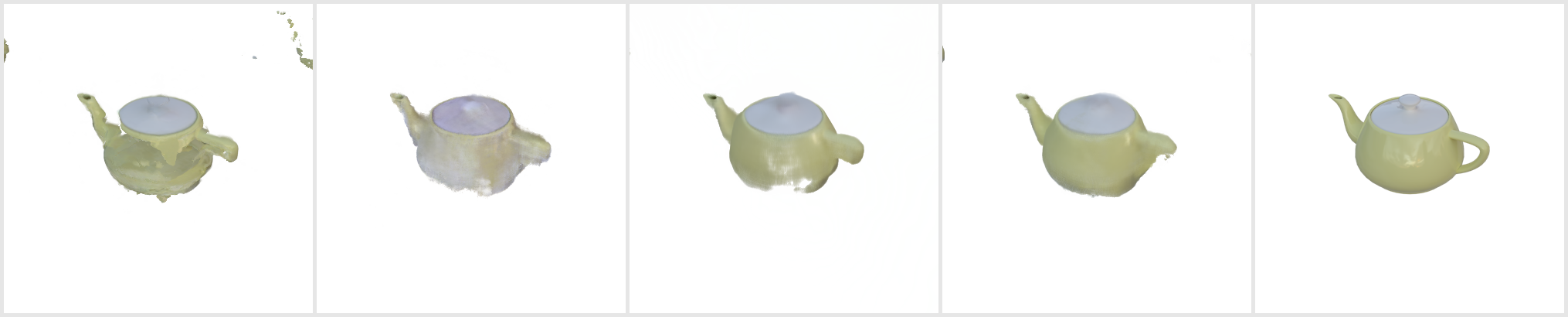}
         \caption{Shiny Blender 4-view}
         \label{fig:shiny}
     \end{subfigure}
\vspace{-2mm}
\caption{\textbf{Qualitative comparisons on Realistic Synthetic 360$^\circ$, DTU, and Shiny Blender.}
Our ARC-NeRF achieves notable rendering quality with better capturing fine details.
$\ddagger$ indicates the W\&B prior.
Additional results are provided in the supplementary material.
}
\vspace{-3mm}
\label{fig:qual_res}
\end{figure*}

\noindent \textbf{Realistic Synthetic 360$^\circ$. \ }
As shown in~\cref{tab:blender} and \cref{fig:qual_res}, our ARC-NeRF achieves the SOTA performance over all the scenarios and metrics.
Compared to RegNeRF and FlipNeRF, which utilize ray augmentation, ours renders superior quality of objects, with better capturing fine details.
Even compared to FreeNeRF, ours shows a clearer texture of target objects, since high-frequency is adaptively regularized, not forcibly masked, during the early training.

\begin{table}[!t]
\centering
\resizebox{0.95\linewidth}{!}{
\begin{tabular}{l|c|c|c|c}
\toprule
& {PSNR $\uparrow$} & {SSIM $\uparrow$} & {LPIPS $\downarrow$} & {Avg. $\downarrow$} \\
  \midrule
 Mip-NeRF~\cite{barron2021mip} & 8.68 & 0.571 & 0.353 & 0.323 \\
 3DGS~\cite{kerbl20233d} & 14.18 & 0.628 & 0.301 & 0.191 \\
\midrule
\multicolumn{5}{l}{\textit{\textbf{Pre-training.}}} \\
\midrule\midrule
PixelNeRF~\cite{yu2021pixelnerf} & 16.82 & 0.695 & 0.270 & 0.147 \\
PixelNeRF$^\dagger$~\cite{yu2021pixelnerf} & 18.95 & 0.710 & 0.269 & 0.125 \\
SRF~\cite{chibane2021stereo} & 15.32 & 0.671 & 0.304 & 0.171 \\
SRF$^\dagger$~\cite{chibane2021stereo} & 15.68 & 0.698 & 0.281 & 0.162 \\
MVSNeRF~\cite{chen2021mvsnerf} & 18.63 & \cellcolor{orange!50}{0.769} & 0.197 & 0.113 \\
MVSNeRF$^\dagger$~\cite{chen2021mvsnerf} & 18.54 & \cellcolor{orange!50}{0.769} & 0.197 & 0.113 \\
\midrule
\multicolumn{5}{l}{\textit{\textbf{Regularization.}}} \\
\midrule\midrule
DietNeRF~\cite{jain2021putting} & 11.85 & 0.633 & 0.314 & 0.243 \\
RegNeRF~\cite{niemeyer2022regnerf} & 18.89 & 0.745 & 0.190 & 0.112 \\
MixNeRF~\cite{seo2023mixnerf} & 18.95 & 0.744 & 0.203 & 0.113 \\
SimpleNeRF~\cite{somraj2023simplenerf} & 16.25 & 0.751 & 0.249 & 0.143 \\
DiffusioNeRF~\cite{wynn2023diffusionerf} & 16.20 & 0.698 & \cellcolor{yellow!50}{0.160} & 0.128 \\
SparseNeRF~\cite{wang2023sparsenerf} & \cellcolor{orange!50}{19.55} & \cellcolor{orange!50}{0.769} & 0.201 & \cellcolor{yellow!50}{0.102} \\
\color{gray}FreeNeRF$^\ddagger$~\cite{yang2023freenerf} & \cellcolor{gray!25}{\color{gray}19.92} & \cellcolor{gray!25}{\color{gray}0.781} & \cellcolor{gray!25}{\color{gray}0.125} & \cellcolor{gray!25}{\color{gray}0.086} \\
FreeNeRF~\cite{yang2023freenerf} & \cellcolor{yellow!50}{19.23} & \cellcolor{orange!50}{0.769} & \cellcolor{orange!50}{0.149} & 0.103 \\
FlipNeRF~\cite{seo2023flipnerf} & \cellcolor{orange!50}{19.55} & \cellcolor{yellow!50}{0.767} & 0.180 & \cellcolor{orange!50}{0.101} \\
SparseGS~\cite{xiong2023sparsegs} & 18.89 & 0.702 & 0.229 & 0.117 \\
\textbf{ARC-NeRF } & \cellcolor{red!50}{\textbf{19.85}} & \cellcolor{red!50}{\textbf{0.773}} & \cellcolor{red!50}{\textbf{0.146}} & \cellcolor{red!50}{\textbf{0.096}} \\
\bottomrule
\end{tabular}}
\vspace{-2mm}
\caption{
    \textbf{Quantitative results on DTU 3-view.}
    Our ARC-NeRF outperforms its baseline FlipNeRF, and generally achieves competitive results.
    Kindly refer to our supplementary material for additional comparison under 6/9-view scenarios.
    $\dagger$ and $\ddagger$ indicate the fine-tuning and the W\&B prior, respectively.
    }
\label{tab:dtu}
\vspace{-5mm}
\end{table}

\noindent \textbf{DTU. \ }
\cref{tab:dtu} shows the quantitative comparisons on DTU.
Our ARC-NeRF outperforms the baseline FlipNeRF, and achieves competitive results overall.
As demonstrated in~\cref{fig:qual_res}, ours shows notable outcomes with sharply rendered textures.
FreeNeRF is trained with the black and white prior assuming the estimated black and white color as the background and table, respectively, which is a highly strong assumption specific to the dataset, and achieves degenerate results without the prior.
However, our ARC-NeRF achieves competitive performance without any heuristic prior by using Area Ray, which enables adaptive regularization of high-frequency.
Kindly refer to the supplementary material for its qualitative comparison.

\begin{table}[!t]
\centering
\resizebox{\linewidth}{!}{
\begin{tabular}{l|c|c|c|c}
\toprule
& {PSNR $\uparrow$} & {SSIM $\uparrow$} & {LPIPS $\downarrow$} & {Average $\downarrow$} \\
\midrule
Ref-NeRF~\cite{verbin2022ref} & \cellcolor{yellow!50}{17.10} & 0.821 & \cellcolor{yellow!50}{0.190} & 0.142 \\
FreeNeRF~\cite{yang2023freenerf} & 16.99 & \cellcolor{yellow!50}{0.828} & \cellcolor{orange!50}{0.157} & \cellcolor{yellow!50}{0.131} \\
FlipNeRF~\cite{seo2023flipnerf} & \cellcolor{orange!50}{18.14} & \cellcolor{orange!50}{0.847} & \cellcolor{red!50}{\textbf{0.141}} & \cellcolor{orange!50}{0.109} \\
\textbf{ARC-NeRF } & \cellcolor{red!50}{\textbf{18.68}} & \cellcolor{red!50}{\textbf{0.851}} & \cellcolor{red!50}{\textbf{0.141}} & \cellcolor{red!50}{\textbf{0.107}} \\
\bottomrule
\end{tabular}
}
\vspace{-2mm}
\caption{\textbf{Quantitative results on Shiny Blender 4-view.}
        While FlipNeRF demonstrates competitive performance by utilizing flipped mirror reflection rays, which are physically aligned with the non-Lambertian surfaces, our approach consistently outperforms FlipNeRF and other comparative methods.
        }
\label{tab:shiny}
\vspace{-5mm}
\end{table}

\noindent \textbf{Shiny Blender. \ }
We additionally compare our ARC-NeRF and others on Shiny Blender, which is largely non-Lambertian, as in~\cref{fig:shiny} and~\cref{tab:shiny}.
Although FlipNeRF achieves competitive performance thanks to its ray augmentation utilizing mirror reflection, which accords with the non-Lambertian surfaces, ours still outperforms others including FlipNeRF.
We conjecture that under the few-shot setting where the major challenge is to learn 3D geometry effectively while preventing the overfitting, our adaptive high-frequency regularization plays an important role.

\section{Conclusion}
\label{sec:conclusion}

In this work, we have approached the few-shot novel view synthesis from the perspective of a ray parameterization.
Our ARC-NeRF casts an Area Ray as an augmented ray, which effectively prevents overfitting by adaptively regularizing the high-frequency components based on the target pixel photo-consistency.
Furthermore, as a single Area Ray covers a broader range of unseen view areas than an original ray, our ARC-NeRF is trained more efficiently without casting multiple augmented rays per original ray, maximizing the use of ray augmentation.
We also utilize a free lunch training resource, \ie luminance map, to provide additional supervision for output densities, further improving the fine details.
Our ARC-NeRF achieves the SOTA or competitive performance compared to other methods under various few-shot scenarios.
We expect that our ARC-NeRF is able to inspire a meaningful approach in the few-shot NeRF.

% \paragraph{Acknowledgements.}
\noindent \textbf{Acknowledgements. \ }
This work was supported by NRF grant (2021R1A2C3006659) and IITP grant (RS-2021-II211343), both funded by MSIT of the Korean Government.
The work was also supported by Samsung Electronics (IO201223-08260-01).
{
    \small
    \bibliographystyle{ieeenat_fullname}
    \bibliography{main}
}

\clearpage
\setcounter{page}{1}
\setcounter{section}{0}
\setcounter{table}{0}
\setcounter{figure}{0}
\renewcommand\thesection{\Alph{section}}
\renewcommand\thetable{\Alph{table}}
\renewcommand\thefigure{\Alph{figure}}

\twocolumn[{
\renewcommand\twocolumn[1][]{#1}
\maketitlesupplementary
\centering
\resizebox{0.9\linewidth}{!}{
\begin{tabular}{l|c|c|c|c|c|c}
\toprule
  Hyperparameter & \multicolumn{2}{c|}{Realistic Synthetic 360$^\circ$~\cite{mildenhall2021nerf}} & \multicolumn{3}{c|}{DTU~\cite{jensen2014large}} & Shiny Blender~\cite{verbin2022ref} \\
  \& Balancing Weights & 4-view & 8-view & 3-view & 6-view & 9-view & 4-view \\
  \midrule
  LR & \multicolumn{2}{c|}{$[1\mathrm{e}{-3}, 1\mathrm{e}{-5}]$} & \multicolumn{3}{c|}{$[2\mathrm{e}{-3}, 2\mathrm{e}{-5}]$} & $[1\mathrm{e}{-3}, 1\mathrm{e}{-5}]$ \\
  \midrule
  Warm-up Iter. & 512 & 1024 & 512 & \multicolumn{2}{c|}{2048} & 512 \\
  \midrule
  $\eta_{\text{Ori.}} (\text{for }\mathcal{L}_\text{Ori.})$ & $1\mathrm{e}{-1}$ & $1\mathrm{e}{-2}$ & $1\mathrm{e}{-1}$ & \multicolumn{2}{c|}{$1\mathrm{e}{-2}$} & $1\mathrm{e}{-1}$ \\
  \midrule
  $\eta_{\text{lum.}} (\text{for }l_\text{lum.})$ & $1\mathrm{e}{-3}$ & $1\mathrm{e}{-4}$ & $1\mathrm{e}{-3}$ & $1\mathrm{e}{-4}$ & $1\mathrm{e}{-5}$ & $1\mathrm{e}{-3}$ \\
  \midrule
  $\tilde{\eta}_{\text{lum.}} (\text{for }\tilde{l}_\text{lum.})$ & $1\mathrm{e}{-4}$ & $1\mathrm{e}{-5}$ & $1\mathrm{e}{-4}$ & $1\mathrm{e}{-5}$ & $1\mathrm{e}{-6}$ & $1\mathrm{e}{-4}$ \\
\bottomrule
\end{tabular}}
\vspace{-2mm}
\captionof{table}{
    \textbf{Hyperparameters and balancing weights.}
    Since our ARC-NeRF is built upon FlipNeRF, we follow the training details for other hyperparameters, which are not mentioned here, as FlipNeRF.
    $[\alpha, \beta]$ denotes the annealing from $\alpha$ to $\beta$.
    }
    \label{supp_tab:hyperparameters}
\vspace{5mm}
}]

%%%%%%%%% BODY TEXT
\section{Experimental Setting}
\label{supp_sec:experimental_setting}

\begin{figure}[t]
\centering
\includegraphics[width=0.8\linewidth]{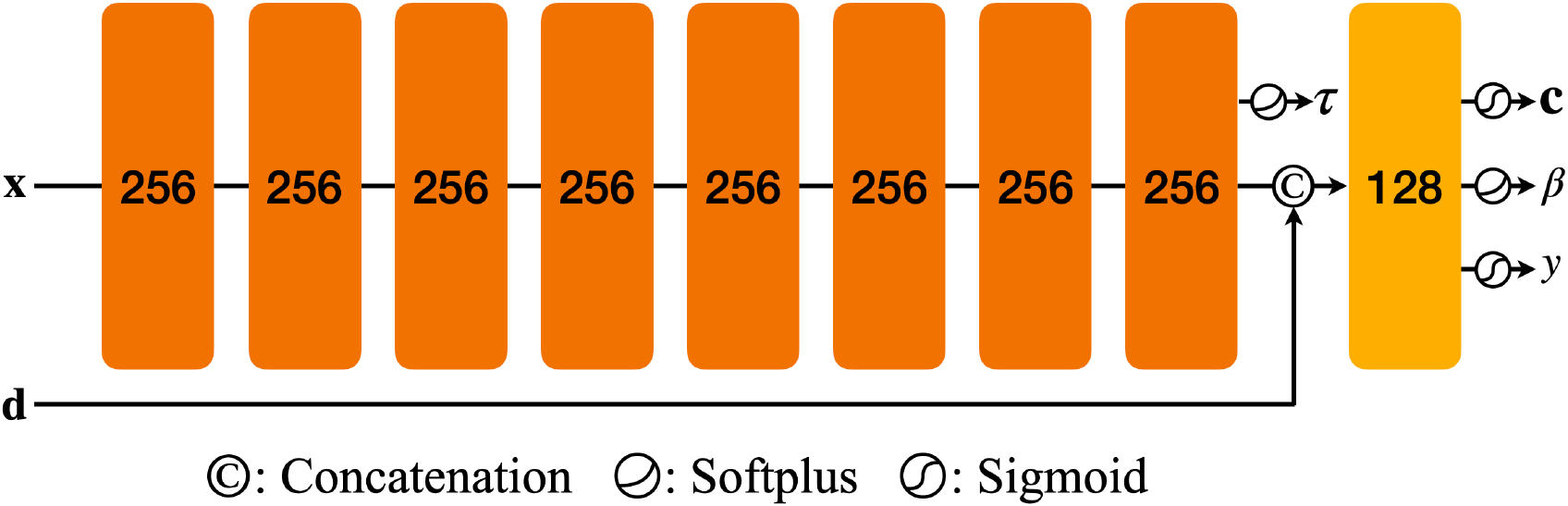}
\caption{\textbf{Network architecture of ARC-NeRF.}
        Our ARC-NeRF estimates the additional output $y$, \ie the relative luminance for our auxiliary luminance estimation task.
        }
\vspace{-3mm}
\label{supp_fig:arch}
\end{figure}

\paragraph{\textbf{Implementational details.}}
Our ARC-NeRF is implemented upon FlipNeRF~\cite{seo2023flipnerf}, and we follow its overall training scheme.
We utilize the scene space annealing strategy during the initial training phase following~\cite{niemeyer2022regnerf, seo2023mixnerf, seo2023flipnerf}.
Furthermore, we adopt the initial warm up and exponential decay for the learning rate.
We use the Adam optimizer~\cite{kingma2014adam} with gradient clipping set to 0.1 for both each element of the gradient value and the gradient's norm.
Our ARC-NeRF is trained for 500 pixel epochs using a batch size of 4,096 on four NVIDIA RTX 3090 GPUs.
Additionally, since our proposed Area Ray encompasses broader areas of unseen views compared to a single ray, we set the masking threshold $\psi$ as 45$^\circ$, which is smaller than that of FlipNeRF, to avoid over-regularization effect of augmented rays.
The related experiment is demonstrated in~\cref{supp_tab:exp_mask_threshold}.

\paragraph{\textbf{Hyperparameters.}}
For additional details on hyperparameters and loss balancing terms based on training views and datasets, kindly refer to~\cref{supp_tab:hyperparameters}.
Note that our ARC-NeRF follows the same hyperparameters as FlipNeRF for other training losses and schemes which are not specified in~\cref{supp_tab:hyperparameters}.

\section{Further Details of Method}
\label{supp_sec:implementational_details}
\paragraph{\textbf{Architectural details.}}
Our ARC-NeRF leverages the network architecture of mip-NeRF~\cite{barron2021mip}, which is commonly used in several few-shot NeRF models~\cite{yang2023freenerf, niemeyer2022regnerf, seo2023flipnerf, seo2023mixnerf}.
Moreover, our ARC-NeRF additionally estimates the relative luminance $y$.
Kindly refer to more details in~\cref{supp_fig:arch}.

\paragraph{\textbf{Total loss.}}
Our ARC-NeRF is trained to maximize the log-likelihood of the target pixel $\mathbf{c}_\text{GT}$ for both sets of original input rays $\mathcal{R}$ and our proposed Area Rays $\tilde{\mathcal{R}}$, as well as to minimize the mean squared errors (MSE) between the ground-truth and estimated pixel values.
Except our proposed $\mathcal{L}_\text{lum.}$, we use the same training losses as those of FlipNeRF.
Note that we use $\mathcal{L}_\text{MSE}$ only for $\mathcal{R}$ and exploit a batch of Area Rays instead of flipped reflection rays.
Summing up, the total loss over a batch is calculated as follows:
\begin{equation}
\begin{split}
\mathcal{L}_\text{Total} &= \mathcal{L}_\text{MSE} + \mathcal{L}_\text{lum.} + \eta_{\text{NLL}}\mathcal{L}_\text{NLL} + \tilde{\eta}_{\text{NLL}}\tilde{\mathcal{L}}_\text{NLL} \\
&+ \eta_{\text{UE}}\mathcal{L}_\text{UE} + \tilde{\eta}_{\text{UE}}\tilde{\mathcal{L}}_\text{UE}' + \eta_{\text{BFC}}\mathcal{L}_\text{BFC} + \eta_{\text{Ori.}}\mathcal{L}_\text{Ori.}, \\
&\text{where} \quad \mathcal{L}_\text{lum.} = \eta_\text{lum.}l_\text{lum.} + \tilde{\eta}_\text{lum.}\tilde{l}_\text{lum.}.
\label{supp_eq:total_loss}
\end{split}
\end{equation}
$\eta$'s and $\tilde{\eta}$'s represent the loss balancing weights for the original input rays and additional Area Rays, respectively.

\begin{table}[t]
\centering
\resizebox{\linewidth}{!}{
\begin{tabular}{l|c|c|c|c}
\toprule
 & {PSNR $\uparrow$} & {SSIM $\uparrow$} & {LPIPS $\downarrow$} & {Average Err. $\downarrow$} \\
\midrule
FlipNeRF~\cite{seo2023flipnerf} & 16.47 & 0.866 & 0.091 & 0.095 \\
\midrule
ARC-NeRF & & & & \\
\quad w/o view. jitter. & \underline{16.66} & \underline{0.869} & \underline{0.087} & \underline{0.093} \\
\quad w/ view. jitter. & \textbf{16.86} & \textbf{0.873} & \textbf{0.084} & \textbf{0.091} \\
\bottomrule
\end{tabular}
}
\vspace{-2mm}
\caption{
\textbf{Effect of Viewing direction jittering.}
On Realistic Synthetic 360$^\circ$ 4-view, we are able to achieve marginal performance improvement while still outperforming FlipNeRF without the jittering strategy.
}
\label{supp_tab:exp_view_jitter}
\end{table}

\begin{table}[!t]
\centering
\resizebox{\linewidth}{!}{
\begin{tabular}{l|c|c|c|c}
\toprule
$\psi$ & {PSNR $\uparrow$} & {SSIM $\uparrow$} & {LPIPS $\downarrow$} & {Average Err. $\downarrow$} \\
\midrule
180$^\circ$ (None) & 18.15 & 0.749 & 0.179 & 0.120 \\
90$^\circ$ & 18.63 & 0.762 & 0.163 & 0.110 \\
75$^\circ$ & \underline{19.02} & 0.764 & 0.156 & \underline{0.105} \\
60$^\circ$ & 18.94 & \underline{0.766} & \underline{0.154} & \underline{0.105} \\
45$^\circ$ & \textbf{19.85} & \textbf{0.773} & \textbf{0.146} & \textbf{0.096} \\
30$^\circ$ & 18.78 & 0.765 & 0.160 & 0.107 \\
15$^\circ$ & 18.65 & 0.761 & 0.163 & 0.111 \\
\bottomrule
\end{tabular}
}
\vspace{-2mm}
\caption{
\textbf{Comparison of masking thresholds.}
Our ARC-NeRF excludes a set of Area Rays, whose angle $\theta$ between the original input ray is over $\psi$, \ie the target pixel photo-consistency is relatively low considering the threshold $\psi$, from a training batch.
$\psi=180^\circ\text{(None)}$ uses a whole batch of newly generated Area Rays.
}
\label{supp_tab:exp_mask_threshold}
\end{table}

\section{Additional Experiments}
\label{supp_sec:additional_experiments}

\paragraph{\textbf{Viewing direction jittering.}}
For Realistic Synthetic 360$^\circ$~\cite{mildenhall2021nerf} and Shiny Blender~\cite{verbin2022ref}, which consist of inward-facing synthetic scenes with objects located at the center, we adopt the viewing direction jittering, which is a minor additional strategy slightly improving the performance.
We simply add the Gaussian random noise to the input viewing direction $\mathbf{d}$ to improve the robustness for the slight change of viewpoints.
As shown in~\cref{supp_tab:exp_view_jitter}, we are able to achieve marginal improvement of rendering quality while still outperforming its baseline, FlipNeRF, even without the jittering strategy.

\paragraph{\textbf{Masking thresholds.}}
Our ARC-NeRF utilizes an additional batch of Area Rays covering a broader area of unseen views, and the high-frequency components of samples along an Area Ray are adaptively regularized via Integrated Positional Encoding (IPE) depending on the angle between the original input direction and the estimated normal vector, \ie the target pixel photo-consistency.
As a result, with the same $\psi=90^\circ$ as FlipNeRF, our ARC-NeRF might suffer from the performance degradation due to over-regularization.
As demonstrated in~\cref{supp_tab:exp_mask_threshold}, our ARC-NeRF achieves the best result with $\psi=45^\circ$.
The larger $\psi$ becomes than $45^\circ$, the worse the performance, as an Area Ray covering too wide area of unseen views leads to over-regularization, which adversely affects the training.
On the other hand, a smaller $\psi$ than $45^\circ$ also leads to poorer performance, as the newly generated Area Rays are excessively filtered, resulting in only a limited number of augmented Area Rays being utilized for training.
Note that the masking threshold $\psi$ depends on the characteristics of casting ray rather than being the hyperparameter which needs to be finetuned elaboratively.

\begin{figure}[!t]
\centering
\resizebox{\linewidth}{!}{
\begin{tabular}{l|c|c|c|c}
\toprule
& {PSNR $\uparrow$} & {SSIM $\uparrow$} & {LPIPS $\downarrow$} & {Average $\downarrow$} \\
\midrule
FreeNeRF$^\ddagger$~\cite{yang2023freenerf} & \textbf{19.92} & \textbf{0.781} & \textbf{0.125} & \textbf{0.086} \\
FreeNeRF~\cite{yang2023freenerf} & 19.23 & 0.769 & 0.149 & 0.103 \\
\textbf{ARC-NeRF }  & \underline{19.85} & \underline{0.773} & \underline{0.146} & \underline{0.096} \\
\bottomrule
\end{tabular}}
\resizebox{\linewidth}{!}{
\begin{subfigure}[b]{\linewidth}
\vspace{1mm}
\begin{tabular}{p{0.3\textwidth}p{0.27\textwidth}p{0.3\textwidth}}
     \centering\scriptsize FreeNeRF$^\ddagger$~\cite{yang2023freenerf} & \centering\scriptsize FreeNeRF~\cite{yang2023freenerf} & \centering\scriptsize \textbf{ARC-NeRF}
\end{tabular}
\includegraphics[width=\linewidth]{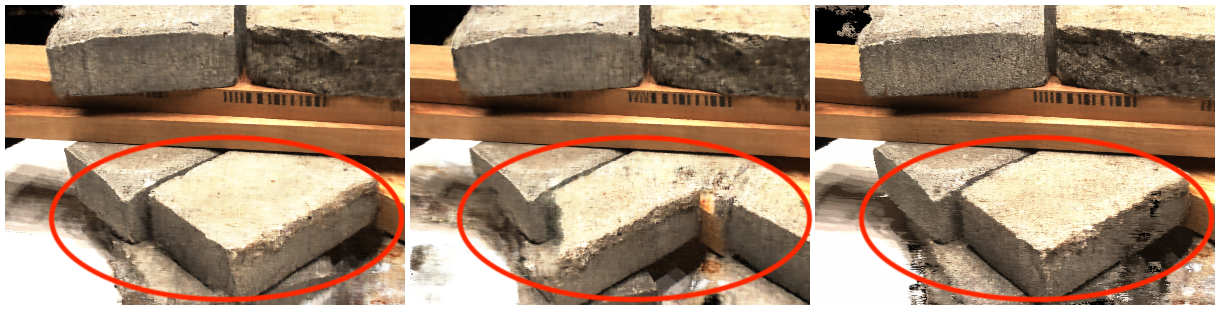}
\end{subfigure}
}
\vspace{-5mm}
\caption{\textbf{Quantitative and qualitative comparison with FreeNeRF on DTU 3-view.}
        Although FreeNeRF achieves high-quality of rendering with only a few images, it depends on the white and black prior, which is highly heuristic based on the characteristics of dataset.
        Our ARC-NeRF achieves comparable performance to FreeNeRF without the heuristic prior for training.
        $\ddagger$ denotes the W\&B prior.
        }
\label{supp_fig:exp_wbprior}
\end{figure}

\begin{table*}[!t]
\centering
\resizebox{\linewidth}{!}{
\begin{tabular}{l|c|ccc|ccc|ccc|ccc}
\toprule
  & \multirow{2}{*}{Method} & \multicolumn{3}{c}{PSNR $\uparrow$} & \multicolumn{3}{c}{SSIM $\uparrow$} & \multicolumn{3}{c}{LPIPS $\downarrow$} & \multicolumn{3}{c}{Avg. Err. $\downarrow$}  \\
  & & 3-view & 6-view & 9-view  & 3-view & 6-view & 9-view  & 3-view & 6-view & 9-view  & 3-view & 6-view & 9-view \\
  \midrule
Mip-NeRF~\cite{barron2021mip} & \multirow{2}{*}{-} & 8.68 & 16.54 & 23.58 & 0.571 & 0.741 & 0.879 & 0.353 & 0.198 & 0.092 & 0.323 & 0.148 & 0.056 \\
3DGS~\cite{kerbl20233d} & & 14.18 & - & - & 0.628 & - & - & 0.301 & - & - & 0.191 & - & - \\
\midrule
PixelNeRF~\cite{yu2021pixelnerf} & \multirow{6}{*}{Pre-training} & 16.82 & 19.11 & 20.40 & 0.695 & 0.745 & 0.768 & 0.270 & 0.232 & 0.220 & 0.147 & 0.115 & 0.100 \\
PixelNeRF$^\dagger$~\cite{yu2021pixelnerf} & & 18.95 & 20.56 & 21.83 & 0.710 & 0.753 & 0.781 & 0.269 & 0.223 & 0.203 & 0.125 & 0.104 & 0.090 \\
SRF~\cite{chibane2021stereo} & & 15.32 & 17.54 & 18.35 & 0.671 & 0.730 & 0.752 & 0.304 & 0.250 & 0.232 & 0.171 & 0.132 & 0.120 \\
SRF$^\dagger$~\cite{chibane2021stereo} & & 15.68 & 18.87 & 20.75 & 0.698 & 0.757 & 0.785 & 0.281 & 0.225 & 0.205 & 0.162 & 0.114 & 0.093 \\
MVSNeRF~\cite{chen2021mvsnerf} & & 18.63 & 20.70 & 22.40 & \cellcolor{orange!50}{0.769} & 0.823 & 0.853 & 0.197 & 0.156 & 0.135 & 0.113 & 0.088 & 0.068 \\
MVSNeRF$^\dagger$~\cite{chen2021mvsnerf} & & 18.54 & 20.49 & 22.22 & \cellcolor{orange!50}{0.769} & 0.822 & 0.853 & 0.197 & 0.155 & 0.135 & 0.113 & 0.089 & 0.069 \\
\midrule
DietNeRF~\cite{jain2021putting} & \multirow{11}{*}{Regularization} & 11.85 & 20.63 & 23.83 & 0.633 & 0.778 & 0.823 & 0.314 & 0.201 & 0.173 & 0.243 & 0.101 & 0.068 \\
RegNeRF~\cite{niemeyer2022regnerf} & & 18.89 & 22.20 & 24.93 & 0.745 & \cellcolor{orange!50}{0.841} & \cellcolor{orange!50}{0.884} & 0.190 & 0.117 & 0.089 & 0.112 & 0.071 & 0.047 \\
MixNeRF~\cite{seo2023mixnerf} & & 18.95 & 22.30 & 25.03 & 0.744 & 0.835 & 0.879 & 0.203 & 0.102 & 0.065 & 0.113 & 0.066 & 0.042 \\
SimpleNeRF~\cite{somraj2023simplenerf} & & 16.25 & 20.60 & 22.75 & 0.751 & 0.828 & 0.856 & 0.249 & 0.190 & 0.176 & 0.143 & 0.088 & 0.071 \\
DiffusioNeRF~\cite{wynn2023diffusionerf} & & 16.20 & 20.34 & \cellcolor{orange!50}{25.18} & 0.698 & 0.818 & \cellcolor{yellow!50}{0.883} & \cellcolor{yellow!50}{0.160} & \cellcolor{yellow!50}{0.093} & \cellcolor{red!50}{\textbf{0.046}} & 0.128 & 0.072 & \cellcolor{red!50}{\textbf{0.036}} \\
SparseNeRF~\cite{wang2023sparsenerf} & & \cellcolor{orange!50}{19.55} & - & - & \cellcolor{orange!50}{0.769} & - & - & 0.201 & - & - & \cellcolor{yellow!50}{0.102} & - & - \\
\color{gray}FreeNeRF$^\ddagger$~\cite{yang2023freenerf} & & \cellcolor{gray!25}{\color{gray}19.92} & \cellcolor{gray!25}{\color{gray}23.25} & \cellcolor{gray!25}{\color{gray}25.60} & \cellcolor{gray!25}{\color{gray}0.781} & \cellcolor{gray!25}{\color{gray}0.838} & \cellcolor{gray!25}{\color{gray}0.877} & \cellcolor{gray!25}{\color{gray}0.125} & \cellcolor{gray!25}{\color{gray}0.085} & \cellcolor{gray!25}{\color{gray}0.057} & \cellcolor{gray!25}{\color{gray}0.086} & \cellcolor{gray!25}{\color{gray}0.058} & \cellcolor{gray!25}{\color{gray}0.038} \\
FreeNeRF~\cite{yang2023freenerf} & & \cellcolor{yellow!50}{19.23} & \cellcolor{red!50}{\textbf{22.77}} & \cellcolor{red!50}{\textbf{25.59}} & \cellcolor{orange!50}{0.769} & 0.835 & 0.877 & \cellcolor{orange!50}{0.149} & \cellcolor{orange!50}{0.088} & \cellcolor{orange!50}{0.057} & 0.103 & \cellcolor{orange!50}{0.063} & \cellcolor{orange!50}{0.039} \\
FlipNeRF~\cite{seo2023flipnerf} & & \cellcolor{orange!50}{19.55} & \cellcolor{yellow!50}{22.45} & 25.12 & \cellcolor{yellow!50}{0.767} & \cellcolor{yellow!50}{0.839} & 0.882 & 0.180 & 0.098 & \cellcolor{yellow!50}{0.062} & \cellcolor{orange!50}{0.101} & \cellcolor{yellow!50}{0.064} & 0.041 \\
SparseGS~\cite{xiong2023sparsegs} & & 18.89 & - & - & 0.702 & - & - & 0.229 & - & - & 0.117 & - & - \\
\textbf{ARC-NeRF } & & \cellcolor{red!50}{\textbf{19.85}} & \cellcolor{orange!50}{22.73} & \cellcolor{yellow!50}{25.14} & \cellcolor{red!50}{\textbf{0.773}} & \cellcolor{red!50}{\textbf{0.842}} & \cellcolor{red!50}{\textbf{0.886}} & \cellcolor{red!50}{\textbf{0.146}} & \cellcolor{red!50}{\textbf{0.084}} & \cellcolor{orange!50}{0.057} & \cellcolor{red!50}{\textbf{0.096}} & \cellcolor{red!50}{\textbf{0.060}} & \cellcolor{yellow!50}{0.040} \\
\bottomrule
\end{tabular}
}
\caption{
    \textbf{Additional quantitative comparison on DTU.}
    Our ARC-NeRF shows competitive performance by outperforming other methods on most metrics, even without relying on any dataset-specific priors.
    $\dagger$ and $\ddagger$ indicate fine-tuning and W\&B prior, respectively.
    }
    \label{supp_tab:dtu}
\end{table*}

\begin{figure*}[!t]
\centering
     \begin{tabular}{p{0.02\textwidth}p{0.128\textwidth}p{0.128\textwidth}p{0.128\textwidth}p{0.128\textwidth}p{0.128\textwidth}p{0.128\textwidth}p{0.03\textwidth}}
     & \centering\scriptsize mip-NeRF~\cite{barron2021mip} & \centering\scriptsize MixNeRF~\cite{seo2023mixnerf} & \centering\scriptsize  FreeNeRF~\cite{yang2023freenerf} & \centering\scriptsize FlipNeRF~\cite{seo2023flipnerf} & \centering\scriptsize \textbf{ARC-NeRF} & \centering\scriptsize Ground Truth &
    \end{tabular}
\begin{subfigure}[b]{0.92\textwidth}
         \centering
        \includegraphics[width=\linewidth]{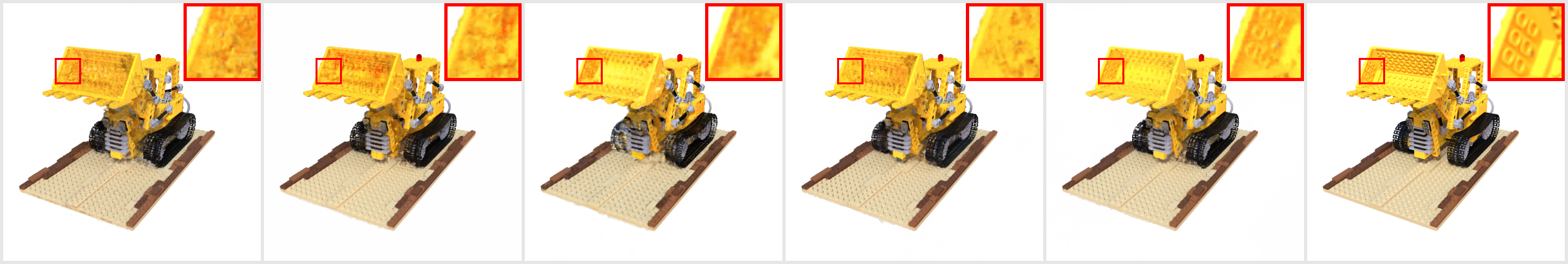}
         \caption{Realistic Synthetic 360$^\circ$ 8-view}
         \label{supp_fig:qual_res_blender_8view}
     \end{subfigure}
    \begin{tabular}{p{0.02\textwidth}p{0.128\textwidth}p{0.128\textwidth}p{0.128\textwidth}p{0.128\textwidth}p{0.128\textwidth}p{0.128\textwidth}p{0.03\textwidth}}
     & \centering\scriptsize mip-NeRF~\cite{barron2021mip} & \centering\scriptsize RegNeRF~\cite{niemeyer2022regnerf} & \centering\scriptsize  FreeNeRF$^\ddagger$~\cite{yang2023freenerf} & \centering\scriptsize FlipNeRF~\cite{seo2023flipnerf} & \centering\scriptsize \textbf{ARC-NeRF} & \centering\scriptsize Ground Truth &
    \end{tabular}
\begin{subfigure}[b]{0.92\textwidth}
         \centering
        \includegraphics[width=\linewidth]{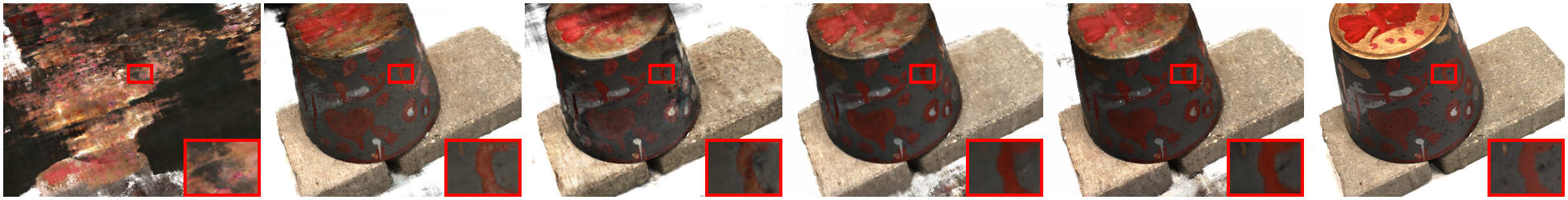}
         \caption{DTU 6-view}
         \label{supp_fig:qual_res_dtu_6view}
     \end{subfigure}
\begin{subfigure}[b]{0.92\textwidth}
         \centering
        \includegraphics[width=\linewidth]{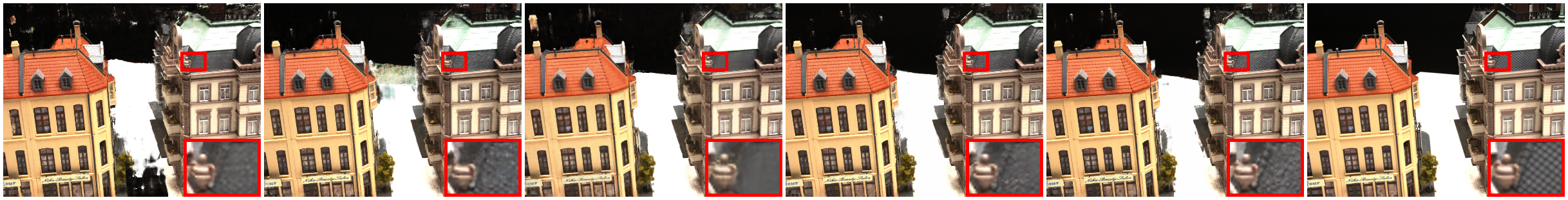}
         \caption{DTU 9-view}
         \label{supp_fig:qual_res_dtu_9view}
     \end{subfigure}
\vspace{-2mm}
\caption{\textbf{Additional qualitative comparisons.}
Our ARC-NeRF achieves high-quality renderings with fine details and clearer texture.}
\vspace{-3mm}
\label{supp_fig:qual_res}
\end{figure*}

\paragraph{\textbf{Additional results.}}
\cref{supp_fig:exp_wbprior} shows the comparison with our ARC-NeRF, FreeNeRF~\cite{yang2023freenerf}, and FreeNeRF without the white and black prior.
While FreeNeRF achieves high-quality rendering, it relies on a white and black prior, \ie a highly heuristic approach considering the specific dataset's characteristics.
In contrast, our ARC-NeRF outperforms FreeNeRF without relying on such priors, even showing competitive performance compared to FreeNeRF with the prior.

The quantitative comparisons including the 6/9-view scenarios on DTU and more qualitative results are demonstrated in~\cref{supp_tab:dtu} and \cref{supp_fig:qual_res}, respectively.
Note that we report the results of other methods from their original papers or~\cite{niemeyer2022regnerf, seo2023mixnerf, seo2023flipnerf}, which outperformed the results from the corresponding original paper by modified training curriculum~\cite{niemeyer2022regnerf}.
Our ARC-NeRF achieves competitive performance among the SOTA methods.
Furthermore, our supplementary videos show the comparison with other methods on Realistic Synthetic 360$^{\circ}$ and FreeNeRF on DTU.
% Kindly note that we were unable to include the video results due to file format and size limitations, but they will be made available upon paper acceptance.

\section{Limitations and Future Work}
\label{supp_sec:limitation}
Our proposed ARC-NeRF excels at capturing finer textures and details of object surfaces but shows limitations when addressing complex backgrounds, such as unbounded scenes composed of widely varying depths.
This is due to potential obstacles between cast rays, which hinder pixel photo-consistency.
Developing a new ray parameterization that can cover broad unseen view areas across significantly varied depths while effectively dealing with the obstacles would be a meaningful extension toward few-shot view synthesis for unbounded scenes as a future work.

\newpage
% {
    % \small
    % \bibliographystyle{ieeenat_fullname}
    % \bibliography{main}
% }

\end{document}